\documentclass{article}
\PassOptionsToPackage{numbers, compress}{natbib}
\usepackage[main, final]{neurips_2026}

\usepackage[utf8]{inputenc}
\usepackage[T1]{fontenc}
\usepackage{hyperref}
\usepackage{url}
\usepackage{booktabs}
\usepackage{amsfonts}
\usepackage{amsmath}
\usepackage{amssymb}
\usepackage{nicefrac}
\usepackage{microtype}
\usepackage{xcolor}
\usepackage{graphicx}
\usepackage[normalem]{ulem}
\usepackage{algorithm}
\usepackage{algpseudocode}
\usepackage{booktabs}
\usepackage{multirow}
\usepackage{graphicx}
\usepackage{array}
\usepackage{adjustbox}
\newcommand{\xmark}{\ensuremath{\times}}
\newcommand{\rng}[2]{\begin{tabular}[c]{@{}c@{}}$\in[#1]$\\[-1pt]{\tiny obs.\ #2}\end{tabular}}
\newcommand{\method}{\textsc{Discover}}

\title{\textsc{Discover}: Online Variance-Guided Data Discovery for Budgeted Multimodal GRPO}

\author{%
  \makebox[\textwidth][c]{%
    \begin{minipage}[t]{0.29\textwidth}
      \centering
      \normalfont\normalsize
      {\bfseries Peyman Gholami}\\
      {\small University of British Columbia}\\
      {\small Vector Institute for AI}\\
      {\footnotesize\texttt{peymang@cs.ubc.ca}}
    \end{minipage}%
    \hfill
    \begin{minipage}[t]{0.29\textwidth}
      \centering
      \normalfont\normalsize
      {\bfseries Shivam Chandhok$^\ast$}\\
      {\small University of British Columbia}\\
      {\small Vector Institute for AI}\\
      {\footnotesize\texttt{chshivam@cs.ubc.ca}}
    \end{minipage}%
    \hfill
    \begin{minipage}[t]{0.38\textwidth}
      \centering
      \normalfont\normalsize
      {\bfseries Saber Malekmohammadi\thanks{Equal contribution}}\\
            {\small Stanford University,}\\
      {\small UTHealth Houston}\\
      {\footnotesize\texttt{sabermm@stanford.edu}}
    \end{minipage}%
  }
  \AND
  \makebox[\textwidth][c]{%
    \begin{minipage}[t]{0.29\textwidth}
      \centering
      \normalfont\normalsize
      {\bfseries Shayan Shams}\\[0.5\baselineskip]
      {\small UTHealth Houston}\\
      {\footnotesize\texttt{shayan.shams@uth.tmc.edu}}
    \end{minipage}%
    \hfill
    \begin{minipage}[t]{0.29\textwidth}
      \centering
      \normalfont\normalsize
      {\bfseries Robert Xiao}\\[0.5\baselineskip]
      {\small University of British Columbia}\\
      {\footnotesize\texttt{brx@cs.ubc.ca}}
    \end{minipage}%
    \hfill
    \begin{minipage}[t]{0.38\textwidth}
      \centering
      \normalfont\normalsize
      {\bfseries Leonid Sigal}\\
      {\small University of British Columbia,}\\
      {\small\mbox{Vector Institute for AI, CIFAR AI Chair}}\\
      {\footnotesize\texttt{lsigal@cs.ubc.ca}}
    \end{minipage}%
  }
  \AND
  \raisebox{12pt}{%
    \hypersetup{hidelinks}%
    \color{blue!40}%
    \url{https://peymoon.github.io/discover/}%
  }
}

\begin{document}

\maketitle

\vspace{-10mm}

\begin{abstract}
Recent advances in reinforcement learning have made verifiable rewards a central
post-training paradigm for improving multimodal reasoning. However,
methods such as GRPO remain data- and compute-inefficient, spending
rollout and annotation budget on prompts that provide little learning signal.
We propose \textsc{Discover}, an online data-discovery framework for budgeted
multimodal GRPO that enables vision-language models to dynamically identify
prompts that are most useful for the current policy. \textsc{Discover} first organizes
the multimodal sample pool into semantic clusters and queries a small set of
density-diverse representative samples to obtain initial reward observations.
During training, it estimates the utility of unannotated samples by transferring
sparse reward feedback from queried samples. Since sample utility changes as the
policy evolves, \textsc{Discover} dynamically updates these estimates with
recency-weighted rollout observations. It then selects only high-utility prompts
for annotation using a policy-conditioned score that accounts for reward variance and 
difficulty. Once the
discovery budget is exhausted, \textsc{Discover} continues with utility-based replay. Experiments on ViRL39K 
and VLAA-Thinking with
Qwen3-VL-2B show that \textsc{Discover} substantially improves multimodal mathematical
reasoning under a strict 10\% unique-sample budget, outperforming
budget-matched random selection, hard-example sampling, full-pool GRPO under the
same training-step budget, and online selection baselines. Ablations further
validate the importance of semantic coverage, online reward transfer, 
and stable discovery cadence, establishing \textsc{Discover} as a practical
framework for data-efficient multimodal RL post-training.
\end{abstract}

\section{Introduction}
\label{sec:intro}

% Reinforcement learning with verifiable rewards (RLVR) has become the
% dominant post-training paradigm for reasoning-capable language (LLM) and
% vision-language (VLM) models. 
% Group-relative policy optimization (GRPO)
% \citep{shao2024deepseekmath}, which replaces PPO's learned critic
% \citep{schulman2017ppo} with within-group reward comparisons across $G$
% rollouts of a single prompt, has became one of the dominant algorithms for RLVR in recent years. GRPO has led to strong gains on mathematical and visual reasoning benchmarks for both text-only 
% \citep{guo2025deepseekr1,liu2025drgrpo,yu2025dapo} and vision-language
% models \citep{huang2025visionr1,wang2025virl,zhang2025r1vl}. 

%Saber:
Reinforcement learning with verifiable rewards (RLVR) is the leading post-training paradigm for reasoning in large language and vision-language models (LLMs and LVLMs). A primary algorithm is Group-Relative Policy Optimization (GRPO) \citep{shao2024deepseekmath}, which replaces the learned critic of PPO \citep{schulman2017ppo} with reward comparisons across $G$ rollouts per prompt. GRPO has significantly improved mathematical and visual reasoning for text-only \citep{guo2025deepseekr1,liu2025drgrpo,yu2025dapo} and vision-language models \citep{huang2025visionr1,wang2025virl,zhang2025r1vl}.

% The primary computational bottleneck in GRPO post-training is the 
% rollout-and-grade pipeline, % requiring 
% \leon{which requires stochastic inference to be run $G$ times on each prompt followed by computation of a reward for each of these rollouts. A major inefficiency intrinsic 
% to GRPO is that prompts yielding identical 
% semantic outcomes produce no policy-gradient 
% signal \citep{wu2026hive,yu2025dapo,zheng2025greso}.In typical datasets many prompts can lead to such degenerate behaviour, where a prompt may either be too easy or too difficult, leading to either all equally good or all equally bad rewards. Further, certain prompts may simply be redundant or nearly redundant. Overall, these observations suggest that  (i) prompts may differ in their learning utility, and (ii) that this utility may change as the model itself evolves in post-training. The inefficiencies that result from uniform sampling of prompts in the traditional GRPO are not only computational in nature either, each prompt also requires human generated annotation (answer) in order to compute the verifiable reward. This may carry an enormous human cost, as for certain problems annotation may require expertise (e.g., a doctor or a  trained mathematician).  
% Thus in practice one may want to bound not only computation, but also the amount of annotation (labeled data)}. 

%Saber
The primary computational bottleneck in GRPO is the rollout-and-grade pipeline requiring $G$ stochastic inferences and rewards per prompt. GRPO is inherently inefficient when prompts yield identical semantic outcomes across rollouts, as these produce no policy-gradient signal \citep{wu2026hive,yu2025dapo,zheng2025greso}. Many prompts are either too easy, too difficult, or redundant, suggesting that learning utility varies by prompt and evolves alongside the model. Beyond this computational overhead, uniform sampling also incurs high human costs for verifiable rewards, which often require expensive expert annotation. In practice, training must account for both computational and annotation budgets.

% \leon{Existing approaches attempt to address these limitations in a few different ways. 
% \textbf{Static and offline data selection} methods, which are typically developed for LLMs, score a fixed pool
% by quality, gradient features, learning impact, or embedding geometry,
% then train on the selected subset
% \citep{evans2024jest,xia2024less,yang2024s2l,liu2024tsds,zhang2024tagcos,liu2024tive,li2025limr,muennighoff2025s1}.
% \textbf{Online prompt selection for RL post-training} methods
% estimate per-prompt difficulty or variance from rollouts and skips or
% up-weights uninformative prompts \citep{sun2025dots,zheng2025greso,shen2025bots,qu2025adarft,qiu2025mopps,baroian2026promptreplay,wu2026hive,gao2025prompt}. 
% \textbf{Multimodal RL post-training} methods adjust optimization 
% signals, replay strategies, or initializations 
% \citep{wang2025virl,zhang2025r1vl,huang2025visionr1}. 
% However, by and large, these methods \emph{do not employ adaptable prompt selection that dynamically caters to the state of the model during post-training, require human (answer) annotation for a full prompt set, or incorporate visual dependence inherent to VLMs.}}

% Saber:
Existing approaches address these through several strategies. \textbf{Static and offline data selection} methods score a fixed pool by quality, gradient features, or embedding geometry and select a subset \citep{evans2024jest,li2025limr,liu2024tsds,liu2024tive,muennighoff2025s1,xia2024less,zhang2024tagcos}. \textbf{Online prompt selection for RL} estimates difficulty or variance from rollouts to skip or up-weight prompts \citep{baroian2026promptreplay,gao2025prompt,qiu2025mopps,shen2025bots,qu2025adarft,sun2025dots,wu2026hive,yang2024s2l,zheng2025greso}. \textbf{Multimodal RL post-training} focuses on adjusting optimization signals, replay strategies, or initializations \citep{huang2025visionr1,wang2025virl,zhang2025r1vl}. We study online prompt selection for multimodal GRPO under a fixed verified-prompt budget. \method{} transfers reward statistics to rank unqueried prompts and allocates the budget accordingly.

Prior to training, \method{} embeds image-question pairs using a multimodal encoder \citep{li2026qwen3vlembedding}, clusters the embedding space, and obtains ground-truth rewards for density-diverse representatives (REPR) of each cluster. During training, it uses cosine-KNN regression to predict the mean and variance of the reward for the full prompt set and subsequently selects a training subset via variance-guided utility and cluster coverage constraints. These subsets are then added to a time-decayed buffer. \method{} continues to discover new prompts until the annotation budget is exhausted; afterward, it resamples from the frozen pool based on continuously updated utility scores. Thus, the selector functions as a closed-loop controller, allocating the unique-prompt budget and preserving diversity as the policy evolves, while leveraging standard GRPO rollouts as a cost-efficient selection signal.

% \method{} delivers strong empirical performance. We illustrate its benefits by leveraging it when training on two
% post-training datasets (ViRL39K \citep{wang2025virl} and VLAA \citep{chen2504vlaa}) under a 10\% unique-prompt budget. We show resulting improved performance on four multimodal
% mathematical reasoning benchmarks -- DynaMath \cite{}, MathVerse \cite{}, MATH-Vision \cite{}, WeMath \cite{}. Specifically, for 
% Qwen3-VL-2B base, we show a $\mathbf{+22.5\%}$ relative improvement over
% the random-$10\%$ baseline at the same training cost. This highlights the importance of data discovery and selection in our method. Detailed ablations shed light on our design, and importance of various components within the proposed method.

%Saber:

\method{} delivers strong empirical performance, which we demonstrate by training on the ViRL39K \citep{wang2025virl} and VLAA-Thinking \citep{chen2504vlaa} datasets under a strict 10\% total verified-prompt budget, counting every unique prompt
that receives verifier feedback, whether it is first queried to initialize the
selector or upon admission to GRPO training. Our approach yields improvements across four multimodal math benchmarks: DynaMath \cite{zou2025dynamath}, MathVerse \cite{zhang2024mathverse}, MATH-Vision \cite{wang2024mathvision}, and WeMath \cite{qiao2025wemath}. Specifically, for Qwen3-VL-2B on ViRL39K, \method{} achieves a $\mathbf{+32.4\%}$ relative gain over a 10\% random baseline under the same optimizer-step budget, averaged over three seeds. These results, supported by detailed ablations, highlight the critical role of our data discovery and selection strategy.

\noindent
{\bf Contributions.} Our contributions can be summarized as follows:  
\begin{enumerate}
    \item \textbf{A policy-conditioned utility score for multimodal GRPO.}
We introduce a score selecting prompts with high GRPO signal utility, combining predicted reward variance, a mean-aware hard-side bias, and an informative reward band (Section~\ref{sec:utility}).

    \item \textbf{Efficient sparse-to-dense score transfer.}
    % Computing the full score for every prompt would require dense rollouts and
    % dense counterfactual visual probes. Instead, \method{} observes reward
    % statistics and \vll{} measurements only on a small online subset, stores
    % them in recency-weighted buffers, and transfers them to the full prompt
    % pool with cosine-KNN regression over frozen multimodal embeddings
    % (Sections~\ref{sec:observations}--\ref{sec:utility}).
    %\Saber:
    To avoid the costs of dense rollouts and visual probes, \method{} measures reward statistics on a small online subset. These are stored in recency-weighted buffers and transferred to the full prompt pool via cosine-KNN regression over frozen multimodal embeddings (Section~\ref{sec:observations}).

    \item \textbf{Data-efficient multimodal GRPO evidence.}
    % We show that this score-and-transfer strategy improves multimodal
    % mathematical reasoning under a fixed unique-prompt budget. We use
    % ablations to isolate the roles of coverage, online score transfer, visual
    % dependence, and cadence stability. The slow/fast schedule comparison is
    % treated as validation of the scoring mechanism, not as a separate
    % contribution (Section~\ref{sec:experiments}).
    % Saber:
    This score-and-transfer strategy improves multimodal mathematical reasoning under a fixed unique-prompt budget. Through ablations, we isolate the roles of coverage, online score transfer, utility shaping, and cadence stability. A held-out audit verifies that the gains also transfer to image-dependent prompts (Section~\ref{sec:experiments}).
\end{enumerate}

\section{Related Work}
\label{sec:related}

\paragraph{RL post-training and GRPO.}
% GRPO and its variants \citep{shao2024deepseekmath,guo2025deepseekr1,liu2025drgrpo,yu2025dapo} adapt PPO \citep{schulman2017ppo} to verifiable rewards by replacing the learned critic with group-relative normalization. Recent theory shows sample efficiency for GRPO peaks in a balanced-difficulty regime, governed by a relative budget \citep{wachi2026relativebudget}. Since the budget shifts as the policy (model) improves, these ideal training prompts are strictly non-stationary. \method{} tracks this moving target via recency-weighted estimation, dynamically curating the active prompt set under a fixed unique-prompt budget without altering the underlying GRPO objective.

% Saber:
GRPO and its variants \citep{shao2024deepseekmath,guo2025deepseekr1,liu2025drgrpo,yu2025dapo} adapt PPO \citep{schulman2017ppo} to verifiable rewards using group-relative normalization instead of a learned critic. Theory suggests GRPO efficiency peaks in a balanced-difficulty regime determined by a relative budget \citep{wachi2026relativebudget}; as the policy (model) improves, the ideal training prompts are non-stationary. \method{} tracks this moving target via recency-weighted estimation, dynamically curating the active set under a fixed unique-prompt budget without modifying the GRPO objective.

\paragraph{Online prompt selection for RL post-training.}
% A growing literature scores prompts adaptively during RL fine-tuning,
% using reward variance or difficulty filtering
% \citep{sun2025dots,zheng2025greso,qu2025adarft}, learning-impact estimates
% \citep{li2025limr}, Bayesian success-rate posteriors
% \citep{qiu2025mopps,shen2025bots}, entropy verification
% \citep{wu2026hive}, and on-policy replay or curriculum ranking
% \citep{baroian2026promptreplay,gao2025prompt}. Closest to our setting,
% VADE \citep{hu2025vade} maintains per-sample Beta posteriors and uses
% Thompson sampling with prior decay to select prompts with informative
% current correctness probabilities. Under a strict unique-prompt budget VADE's estimates are cold for
% unqueried prompts and provide no explicit coverage guarantee. MMR1-VAS
% \citep{leng2025mmr1} promotes GRPO reward variance with a score combining
% outcome variance and trajectory diversity, but the trajectory-diversity
% term is available only after generation and does not directly score
% never-queried prompts. \method{} targets this complementary bottleneck by
% deciding which new multimodal prompts should spend scarce verifiable
% feedback, using embedding transfer, cluster coverage, and a separate
% visual-dependence signal. BOTS \citep{shen2025bots} is also related in
% its Bayesian difficulty transfer, but is largely text-focused.

% Saber:
Recent work scores prompts adaptively during RL fine-tuning via reward variance, difficulty filtering \citep{sun2025dots,zheng2025greso,qu2025adarft}, learning impact \citep{li2025limr}, Bayesian posteriors \citep{qiu2025mopps,shen2025bots}, entropy verification \citep{wu2026hive}, or curriculum ranking \citep{baroian2026promptreplay,gao2025prompt}. VADE \citep{hu2025vade} maintains a separate Beta posterior for each prompt, incorporating its rollout outcomes and applying prior decay, and selects prompts through Thompson sampling. MMR1's variance-aware sampling \citep{leng2025mmr1} combines outcome variance with trajectory diversity measured from generated responses. DOTS \citep{sun2025dots} predicts difficulty for unqueried prompts using attention over a small reference set. BOTS \citep{shen2025bots} combines observed outcomes with interpolated estimates from two reference models to update difficulty estimates for selected and unselected tasks. DOTS and BOTS were evaluated on text-only reasoning tasks. \method{} transfers recent reward means and disagreements through frozen multimodal embeddings to rank unverified prompts and allocates the verification budget across semantic clusters under a strict budget on unique verified prompts.

\paragraph{Curricula, active learning, and coresets.}
% Curriculum and self-paced learning~\citep{bengio2009curriculum,kumar2010selfpaced} impose an easy-to-hard ordering under a fixed score, while classical active learning and coreset methods address the complementary problems of label-efficient querying~\citep{settles2009active} and representative subset selection~\citep{sener2018coreset,saran2023vessal}. GePPO formalizes sample reuse for PPO-style updates~\citep{queeney2021geppo}.
% The most related subline is batch active learning, which jointly optimizes diversity and uncertainty within each selected batch. BADGE~\citep{ash2020badge} clusters gradient embeddings via $k$-means$\texttt{++}$, BatchBALD~\citep{kirsch2019batchbald} maximizes batch-level mutual information to suppress redundancy, and Cluster-Margin~\citep{citovsky2021clustermargin} preclusters embeddings before margin scoring to scale to large pools.
% These methods are designed for supervised annotation under a label budget; \method{} instead operates on prompt selection under a rollout budget, where uncertainty is derived from live reward variance across GRPO trajectories and scores are recomputed online at no additional cost.

% Saber:
Curriculum and self-paced learning \citep{bengio2009curriculum,kumar2010selfpaced} use easy-to-hard orderings under a fixed score, while active learning and coresets address label-efficient querying \citep{settles2009active} and representative subset selection \citep{sener2018coreset,saran2023vessal}. Batch active learning methods like BADGE \citep{ash2020badge}, BatchBALD \citep{kirsch2019batchbald}, and Cluster-Margin \citep{citovsky2021clustermargin} jointly optimize diversity and uncertainty within each selected batch for supervised tasks. Unlike these, \method{} selects prompts under a rollout budget, estimating uncertainty from GRPO reward variance and updating scores online at no additional cost.

\paragraph{Multimodal RL post-training.}
VL-Rethinker~\citep{wang2025virl}, R1-VL~\citep{zhang2025r1vl}, and Vision-R1~\citep{huang2025visionr1} modify optimizer signal, replay schedule, or initialization for RL, but treat the prompt pool as a fixed, exchangeable set. Recent optimization-based methods address perception or stability directly: PAPO adds perception-aware regularization \citep{wang2025papo}, MSSR stabilizes single-rollout multimodal RLVR \citep{liu2025mssr}, CalibRL controls exploration with hybrid-policy calibration \citep{huang2026calibrl}, and NoisyRollout diversifies rollouts with distorted images \citep{liu2025noisyrollout}. These methods modify policy updates or rollout generation, while \method{} optimizes scarce feedback allocation and complements existing optimizer-side improvements.

% \paragraph{Visual dependence and shortcut diagnostics.}

% Recent studies show that many multimodal math models can solve problems by relying entirely on text priors, ignoring the provided images altogether \citep{liu2025visualmodality}. This echoes a well-documented history of shortcut learning in VQA \citep{agrawal2018vqacp,hudson2019gqa,shrestha2020negative}. Previous efforts to force models to genuinely process visual information rely on expensive interventions: re-balancing datasets \citep{goyal2017vqav2}, generating artificially perturbed image pairs to penalize blind guessing \citep{chen2020css,gokhale2020mutant,niu2021cfvqa}, or using human-drawn maps to supervise model attention \citep{selvaraju2019hint}. Because these debiasing methods require synthetic image generation, pixel-level labels, or manual human supervision, they are entirely impractical for dynamic training loops. \method{} instead treats visual dependence as a lightweight, selection-time signal. It requires no additional supervision, no image generation, and is cheap enough to evaluate continuously on a small per-round probe set.

\section{Method}
\subsection{Problem Setting and Overall Framework}

We introduce \method{}, an online data-discovery framework for budgeted multimodal GRPO. To train an effective policy while querying verifiable rewards for only a small fraction of the prompt pool $\mathcal{D}=\{x_i\}_{i=1}^{N}$ (where $x_i=(q_i,I_i)$), \method{} employs a structured pipeline: unsupervised semantic clustering, initialization via representative samples, dynamic utility estimation using recency-weighted cosine-KNN transfer, and cluster-balanced discovery transitioning to frozen-pool replay once the budget is exhausted.

\textbf{Problem Setting.}
In the budgeted multimodal GRPO setting, a policy $\pi_{\theta_t}$ is trained at step $t$. For a sample $x_i$, GRPO samples $G$ completions:
\begin{equation}
y_{i,1}, \ldots, y_{i,G} \sim \pi_{\theta_t}(\cdot \mid x_i),
\end{equation}
yielding scalar rewards $\{r^{(t)}_{i,g}\}_{g=1}^G$. The empirical group mean and variance are:
\begin{equation}
\mu_i(t)=\frac{1}{G}\sum_{g=1}^{G} r^{(t)}_{i,g},
\qquad
v_i(t)=\frac{1}{G}\sum_{g=1}^{G}
\left(r^{(t)}_{i,g}-\mu_i(t)\right)^2 .
\label{eq:group-stats}
\end{equation}

\paragraph{Budgeted goal.}
Given the costly nature of verifiable feedback, we constrain the set of unique prompts receiving verifier feedback, $\mathcal{V}_{\le t}$, to a strict budget $B$:
\begin{equation}
|\mathcal{V}_{\le t}| \le B = \lfloor \beta N \rfloor.
\label{eq:budget}
\end{equation}
where $\beta$ is the budget fraction. Within the verified set, $\mathcal{R}_t\subseteq\mathcal{V}_{\le t}$ denotes the prompts used for KNN transfer and $\mathcal{T}_{\le t}\subseteq\mathcal{V}_{\le t}$ the GRPO training pool (Table~\ref{tab:symbols}, Appendix~\ref{app:algorithm}). Reward variance $v_i(t)$ serves as a policy-dependent utility proxy, since prompts with uniform rewards (all correct or all incorrect) yield no group-relative signal. The challenge is to estimate this evolving utility for unqueried prompts while preserving semantic coverage and tracking the current policy.

\textbf{Overall Framework.}
The two-stage pipeline is illustrated in Figure~\ref{fig:pipeline}:
\begin{figure}[t]
\centering
\includegraphics[width=\linewidth]{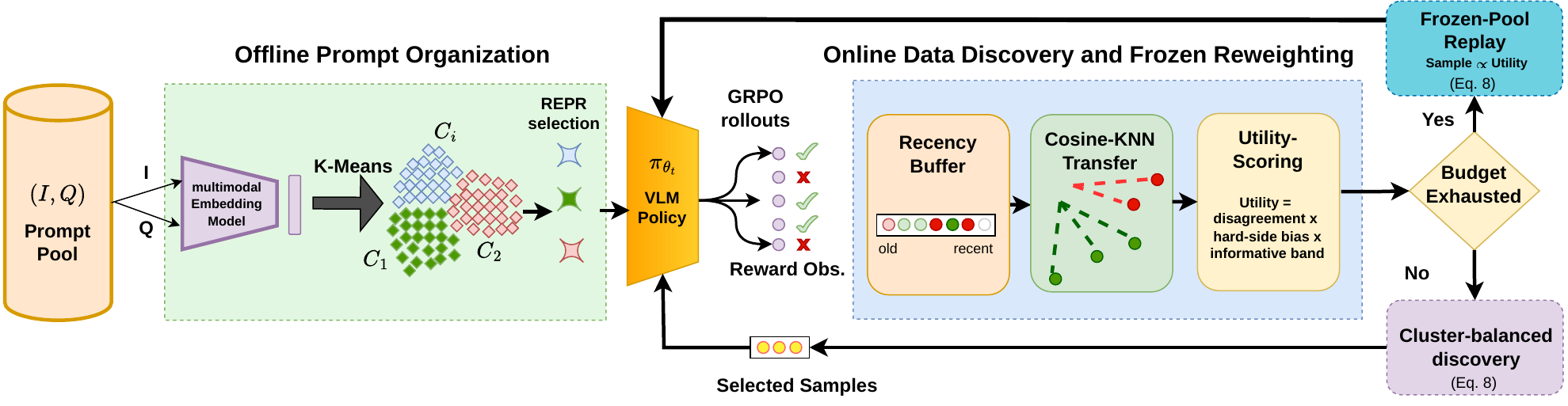}
\vspace{-3mm}
\caption{\textbf{\textsc{Discover} pipeline.} \emph{Offline} (green, left): prompts are embedded, clustered via spherical $k$-means, and seeded with REPR samples. \emph{Online} (blue, right): GRPO rollouts from $\pi_{\theta_t}$ feed a recency-weighted buffer; cosine-KNN transfers sparse statistics to unobserved prompts; utility scoring (Eq.~\ref{eq:discover_utility}) ranks all candidates. The budget switch routes scores to cluster-balanced discovery (Eq.~\ref{eq:cluster}), which admits new prompts until budget $B$ is exhausted, after which frozen-pool replay (Eq.~\ref{eq:reweight}) reweights the existing pool. Both modes return prompts to the policy.\vspace{-0.2in}}
\label{fig:pipeline}
\end{figure}

\begin{enumerate}
    \item \textbf{Offline Multimodal Sample Organization:} The pool is partitioned into semantically coherent clusters using spherical $k$-means on frozen multimodal features. Density-diverse representatives (\textbf{REPR}) are selected to seed initial reward observations and provide neighborhoods for utility transfer.
    \item \textbf{Online Data Discovery and Frozen Reweighting:} During training, \method{} uses rollout feedback to estimate sample utility. Sparse feedback is transferred via cosine-KNN and kept current with recency weighting. High-utility samples are admitted through cluster-balanced discovery until $B$ is reached, after which the pool is frozen and replay is dynamically reweighted.
\end{enumerate}
Initialization evaluates $R\times K$ REPR samples, counting toward budget $B$; they seed the selector and enter GRPO training only through utility-based admission. Newly selected samples refresh the recency-weighted buffer, allowing \method{} to act as a closed-loop controller for data selection.

\subsection{Offline Multimodal Sample Organization}
To ensure semantic diversity, \method{} extracts $L_2$-normalized embeddings $z_i\in\mathbb{R}^{d}$ ($d=2048$) using Qwen3-VL-Embedding-2B. These fixed features support both clustering and cosine-KNN transfer. We apply spherical $k$-means to obtain $K$ clusters, which serve as semantic regions for balanced budget allocation. For unit-norm embeddings, $\|z_i-z_j\|_2^2=2-2z_i^\top z_j$, so squared Euclidean distance decreases monotonically with cosine similarity. Spherical $k$-means maintains unit-norm centroids, assigning prompts by maximum cosine similarity, consistent with cosine-KNN transfer and REPR selection. 
Appendix~\ref{app:embedding-clusters} compares these clusters against standard $k$-means.

To seed the observation buffer, we select up to $R$ REPR samples per
cluster before training. For clusters containing more than $R$ prompts,
candidates are ranked by $g_i=\tilde d_i s_i$, where
$\tilde d_i\in[0,1]$ is the normalized local density computed from
nearest in-cluster neighbours, and $s_i$ is the cosine similarity to
the cluster centroid, clipped to $[0,1]$. We greedily select the
highest-scoring candidate and exclude candidates whose cosine
similarity to it exceeds $0.85$. Clusters containing at most $R$
prompts contribute all their prompts. For reported
configurations, this procedure selects $450$ REPR prompts on
ViRL39K and $400$ on VLAA-Thinking.
Appendix~\ref{app:algorithm} gives the definitions and edge cases.

\subsection{Online Data Discovery and Frozen Reweighting}

\method{} enhances GRPO training through an online discovery loop that estimates reward means and disagreements for all prompts, even when reward observations are sparse. It initializes with representative samples (REPR), maintains statistics via recency weighting, and uses cosine-KNN to infer values for unqueried prompts. After filtering by multimodal utility and selecting new prompts through cluster-balanced discovery, the system freezes the pool once the budget is reached and reweights replay based on the most current utility updates. These steps are explained in the following.

\paragraph{Initialization with REPR samples.}
Training begins by evaluating $R\times K$ representative samples (selected offline) using $G$ rollouts. This seeds reward observations across diverse semantic regions. The initial REPR queries enter the verified set $\mathcal{V}_{\le 0}$ and initialize $\mathcal{R}_0$, counting toward the unique-prompt budget $B$, while the GRPO pool starts empty ($\mathcal{T}_{\le 0}=\emptyset$). REPR prompts remain eligible for GRPO training: their later admission to $\mathcal{T}_{\le t}$ follows the utility rule without increasing the verified-prompt count, since the budget counts unique prompts.

\paragraph{Sparse-to-dense transfer with cosine-KNN.}
\label{sec:observations}
To estimate utility for the majority of unqueried prompts, \method{} transfers statistics from the verified prompts $\mathcal{R}_t\subseteq\mathcal{V}_{\le t}$ with valid recency-weighted statistics using cosine-KNN regression on frozen multimodal features $z$:
\begin{equation}
\hat{y}_i(t)
=
\sum_{j\in \mathrm{NN}_k(i;\mathcal{R}_t)}
\alpha_{ij}(t)y_j,
\qquad
\alpha_{ij}(t)
=
\frac{\exp(\langle z_i,z_j\rangle/T_{\mathrm{nn}})}
{\sum_{\ell\in \mathrm{NN}_k(i;\mathcal{R}_t)}
\exp(\langle z_i,z_\ell\rangle/T_{\mathrm{nn}})} .
\end{equation}
With $k=64$ and $T_{\mathrm{nn}}=0.05$, this nonparametric approach predicts reward means $\hat{\mu}_i(t)$ and disagreement $\hat{v}_i(t)$ without requiring an auxiliary selector network.

\paragraph{Recency-weighted rollout buffer.}
To track the evolving policy, \method{} maintains a buffer where observations at step $\tau$ are discounted by $w(t,\tau)=\exp[-\lambda(t-\tau)]$ ($\lambda=0.05$). Group variance is also normalized by the Bernoulli envelope to isolate usable contrast:
\begin{equation}
\tilde{v}_i(\tau)
=
\operatorname{clip}_{[0,1]}
\left(
\frac{v_i(\tau)}
{\mu_i(\tau)(1-\mu_i(\tau))+\epsilon}
\right).
\end{equation}
For binary correctness rewards, $\tilde v_i(\tau)$ is zero
when all rollouts agree and approximately one otherwise.
Recency weighting and KNN transfer turn this indicator into
a graded estimate of the probability of mixed correctness
outcomes, which provide reward contrast for GRPO. For rewards
bounded in $[m,M]$, the Bhatia--Davis
bound $\mathrm{Var}(r)\le(M-\mu)(\mu-m)$ allows us to replace
$\mu_i(\tau)(1-\mu_i(\tau))$ in the denominator with
$(M-\mu_i(\tau))(\mu_i(\tau)-m)$. Our normalization is the
special case $[m,M]=[0,1]$. To apply this normalization,
unbounded rewards must first be mapped to a bounded range.

We compute the recency-weighted mean and normalized variance as:
\begin{equation}
\bar{\mu}_i(t)
=
\frac{1}{Z_i(t)}
\sum_{\tau\in\Omega_i(t)} w(t,\tau)\mu_i(\tau),
\qquad
\bar{v}_i(t)
=
\frac{1}{Z_i(t)}
\sum_{\tau\in\Omega_i(t)} w(t,\tau)\tilde{v}_i(\tau).
\end{equation}
Here, $\Omega_i(t)$ contains the steps up to $t$ at which rollouts were
recorded for prompt $i$, and $Z_i(t)=\sum_{\tau\in\Omega_i(t)}w(t,\tau)$.
The embedding neighborhoods remain fixed throughout training. At each
selection round, we recompute the reward statistics from the
recency-weighted buffer and transfer them through these neighborhoods to
update utility estimates.

\paragraph{Utility-based filtering.}
\label{sec:utility}
\method{} identifies high-value prompts using a composite utility that balances disagreement with the model’s current capability frontier:
\begin{equation}
u_i(t)
=
\underbrace{\hat{v}_i(t)}_{\text{disagreement}}
\cdot
\underbrace{a(\hat{\mu}_i(t))}_{\text{asymmetry}}
\cdot
\underbrace{m_{\mathrm{dz}}(\hat{\mu}_i(t))}_{\text{dead zone}}.
\label{eq:discover_utility}
\end{equation}
The asymmetry factor
$a(\hat{\mu}_i(t))=1+\eta_h(0.5-\hat{\mu}_i(t))$,
with $\eta_h=0.5$, favors harder prompts at equal predicted
disagreement. The dead-zone filter
$m_{\mathrm{dz}}(\hat{\mu}_i(t))
=\mathbb{I}\{\ell\leq\hat{\mu}_i(t)\leq h\}$,
with $[\ell,h]=[0.05,0.95]$, excludes prompts predicted to be
almost always or almost never solved.

\paragraph{Discovery cadence.}
Discovery slows as the verified-prompt budget fills. Let
$f_t=|\mathcal{V}_{<t}|/B$, where $\mathcal{V}_{<t}$ contains
all prompts verified before discovery at step $t$, including REPR.
Phases $p=1,2,3$ cover $f_t\in[0,0.50)$, $[0.50,0.85)$,
and $[0.85,1)$, respectively. Each phase specifies an admission
fraction $\kappa_p$ and a step interval $\Delta_p$.
At absolute optimizer steps satisfying $t\bmod\Delta_p=0$,
including $t=0$, discovery requests
\[
\bar b_t=\min\!\left\{
\lfloor\kappa_p N_{\rm tr}\rfloor,\,
B-|\mathcal{V}_{<t}|,\,
|\mathcal{U}_t|
\right\},
\qquad
\mathcal{U}_t=\mathcal{D}\setminus\mathcal{V}_{<t},
\]
where $N_{\rm tr}$ is the eligible training-split size.
Once $|\mathcal{V}_{\le t}|=B$, discovery stops permanently
and training continues with frozen-pool replay
(Eq.~\ref{eq:reweight}).
Appendix~\ref{app:selector-config} gives the schedule values.

\paragraph{Cluster-balanced discovery.}
\method{} allocates each round's quota $\bar b_t$ across clusters
according to their mean candidate utility, then admits the
highest-utility prompts within each cluster:
\begin{equation}
\mathcal{A}_t
=
\bigcup_c
\operatorname{Top}_{b_c(t)}(E_c(t);u(t)),
\qquad
E_c(t)=C_c\cap\mathcal{U}_t,
\label{eq:cluster}
\end{equation}
where $\operatorname{Top}_{b_c(t)}$ selects $b_c(t)$ prompts
ranked by utility. For each nonempty candidate set, the cluster
score is
$A_c(t)=|E_c(t)|^{-1}\sum_{i\in E_c(t)}u_i(t)$.
Empty clusters and those with zero mean utility are excluded.
For the remaining clusters, allocation weights are
\[
\rho_c(t)=
\frac{\exp(A_c(t)/T_c)}
{\sum_{c'}\exp(A_{c'}(t)/T_c)},
\qquad T_c=0.5,
\]
where the sum runs over the remaining clusters.
Initial allocations $\mathrm{round}(\bar b_t\rho_c(t))$
are repaired into $b_c(t)$ to respect cluster capacities and meet
the round quota. Any shortfall is filled by adding the highest-utility
unverified prompts overall to $\mathcal{A}_t$. Appendix~\ref{app:algorithm} gives the
repair and tie-breaking rules. Equal-mean clusters receive equal
allocation weight regardless of size, and cluster size enters only
through capacity limits. Selected prompts are verified and
admitted to $\mathcal{T}_{\le t}$.

\paragraph{Frozen-pool reweighting.}
Once the budget is exhausted, the GRPO pool is frozen as the replay pool $\mathcal{P}=\mathcal{T}_{\le t}$; verified prompts never admitted to $\mathcal{P}$ remain Probe-only. Training continues by sampling with replacement from $E_t\subseteq\mathcal{P}$, the prompts that remain inside the mean dead zone, using a utility-proportional distribution and a starvation floor $\epsilon_f=0.05$ to keep prompts eligible:
\begin{equation}
p_i(t)
\propto
\mathbb{I}\{i\in E_t\}
\max\left(
u_i(t),
\epsilon_f \max_{j\in E_t} u_j(t)
\right).
\label{eq:reweight}
\end{equation}
This allows the replay frequency to adapt dynamically as the policy improves.

\begin{figure}[t]
\centering
\includegraphics[width=\linewidth]{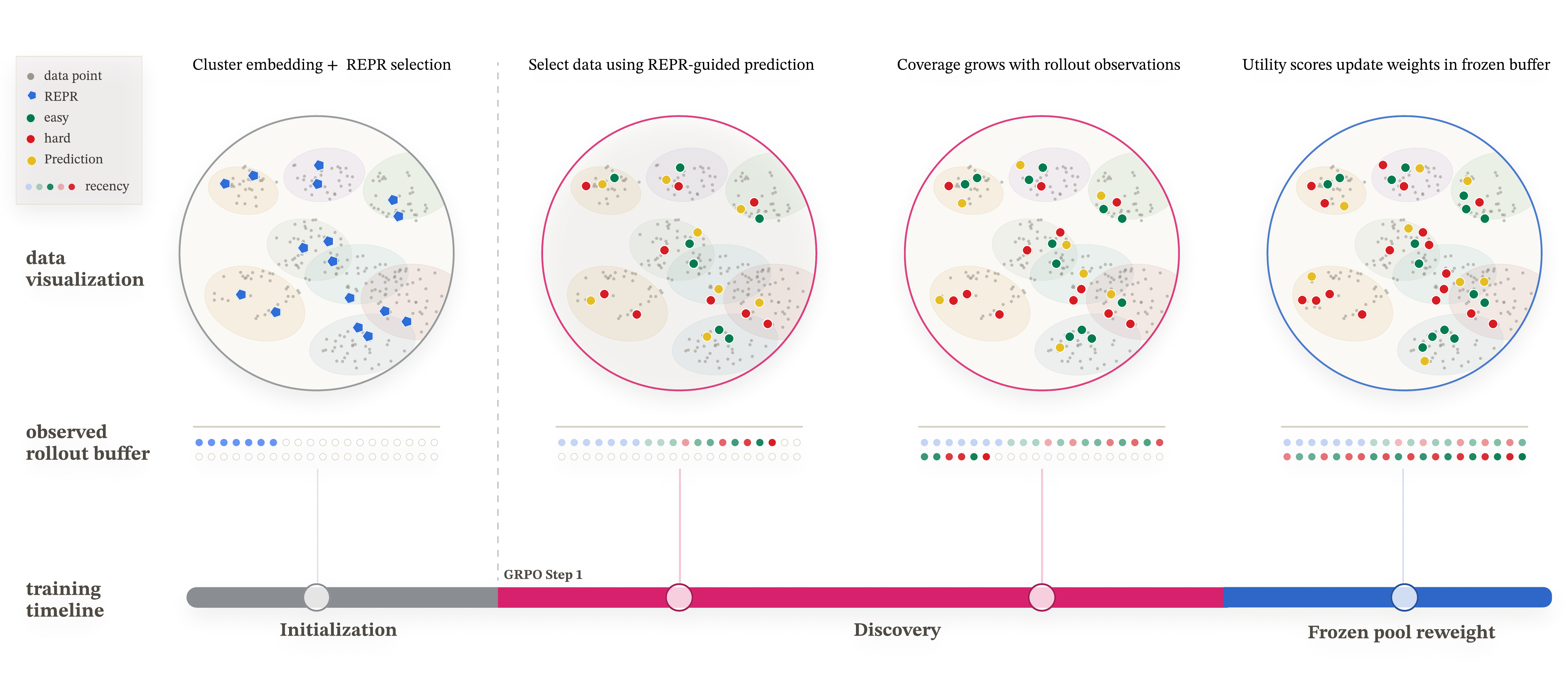}
\vspace{-0.2in}
\caption{\textbf{Cluster-balanced online discovery and frozen-pool reweighting.} \textsc{Discover} builds a fixed multimodal embedding atlas, clusters the prompt pool, and probes REPR to seed a recency-weighted buffer. During discovery, observed easy/hard prompts (green/red) are admitted across clusters while cosine-KNN transfers sparse feedback to unobserved candidates (yellow). Once the unique-prompt budget is exhausted, the pool is frozen and subsequent GRPO steps reweight replay from the buffer; intensity encodes recency.\vspace{-0.2in}}
\label{fig:pool}
\end{figure}

\section{Experiments}
\label{sec:experiments}

\subsection{Setup}
\label{sec:setup}
\paragraph{Training setup.}
We fine-tune Qwen3-VL-2B-Instruct~\citep{bai2025qwen3vl}
with GRPO~\citep{shao2024deepseekmath}, following the standard
training protocol of~\citep{sheng2025hybridflow}, on
ViRL39K~\citep{wang2025virl} and
VLAA-Thinking~\citep{chen2504vlaa}.
All GRPO runs share the training implementation (HybridFlow~\citep{sheng2025hybridflow}),
hardware ($4{\times}$H100), rollout group size ($G{=}8$), exact-match
rule-based verifiers, and optimizer-step budget ($300$ steps on ViRL39K,
$170$ on VLAA-Thinking). GRPO trains on $0.9\times$correctness
$+\,0.1\times$format, with both components binary.
\method{} computes reward statistics in
Eq.~\ref{eq:group-stats} from correctness alone.
Each step budget corresponds to one pass over the eligible training pool
(one epoch), or roughly ten passes over a $10\%$ training pool.
% \paragraph{Training setup.}
% We fine-tune Qwen3-VL-2B-Instruct~\citep{bai2025qwen3vl} with
% GRPO~\citep{shao2024deepseekmath} following standard training protocol from~\citep{sheng2025hybridflow} on two multimodal reasoning datasets :
% ViRL39K~\citep{wang2025virl} and VLAA-Thinking~\citep{chen2504vlaa}.
% Unless otherwise stated, all GRPO-trained rows use the same optimizer-step
% budget, rollout group size, hardware, and training implementation: $300$ steps
% for ViRL39K and $170$ steps for VLAA-Thinking, $G{=}8$ rollouts per prompt,
% $4{\times}$H100 GPUs, HybridFlow~\citep{sheng2025hybridflow}, and exact-match
% rule-based verifiers. GRPO trains on $0.9\times$correctness
% $+\,0.1\times$format, with both components binary. \method{} computes
% the reward means and variances in Eq.~\ref{eq:group-stats} from
% correctness alone. The step counts correspond to one pass over the full
% eligible training pool (i.e one epoch), or roughly ten passes over a $10\%$ budgeted training pool.
% --------------------------------
\paragraph{Budget and exposure accounting.}
We track two costs: annotation cost (unique prompts receiving verifier feedback) and selector-generation cost (\textbf{Probe Gen.}), measured relative to source-pool size $N$. Verified prompts are partitioned by acquisition (\textbf{Selector-first} and \textbf{Training-first}) and final use (\textbf{Train} and \textbf{Probe-only}), with both summing to \textbf{Total verified}. All GRPO runs match optimizer steps, rollout group size, and implementation, differing only in prompt selection and selector overhead. Appendix~\ref{app:prompt-exposure} gives the detailed accounting.

\paragraph{Selector configuration.}
\method{} uses frozen Qwen3-VL-Embedding-2B~\citep{li2026qwen3vlembedding} embeddings with $K{=}150,R{=}3$ on ViRL39K and $K{=}200,R{=}2$ on VLAA-Thinking, combining semantic coverage, cosine-KNN reward-statistic transfer, online refresh, and asymmetric mean-aware utility. In Table~\ref{tab:components}, one-shot denotes a single discovery pass followed by a frozen pool. Clustering and schedule configurations are detailed in Appendix~\ref{app:selector-config}.

\paragraph{Evaluation protocol.}
We track training stability using a held-out $10\%$ validation split and accuracy and actor entropy. For out-of-distribution
generalization, we evaluate checkpoints with VLMEvalKit
\citep{duan2024vlmevalkit} on four external benchmarks covering distinct
multimodal mathematical capabilities: DynaMath \citep{zou2025dynamath}
for visual perturbation robustness, MathVerse
\citep{zhang2024mathverse} for diagram-grounded reasoning, MATH-Vision
\citep{wang2024mathvision} for competition-level problems, and WeMath
\citep{qiao2025wemath} for sub-skill decomposition. We report results with a
512-token generation cap (Table~\ref{tab:main-bench}) and a 2{,}048-token cap (Appendix~\ref{app:eval-2048}).

\paragraph{Baselines.}
We compare \method{} against (i) unmodified Qwen3-VL-2B; (ii) full-data GRPO with matched optimizer steps; (iii) Random-$\beta$, a fixed random subset; (iv) Hard-$\beta$, ranked by base-policy failure rate from a single-rollout over the full training split; (v) MaxVar-$\beta$, ranked by base-policy reward variance from a $G$-rollout scan over full training split; (vi) DOTS-MM, our multimodal adaptation of DOTS~\citep{sun2025dots}; and (vii) VADE-original and VADE-$10\%$~\citep{hu2025vade}, with uncapped and 10\%-capped training pools, respectively. DOTS-MM reference probes count toward the verified-prompt budget; both VADE variants require full-split initialization scans. Baseline details and validation appear in Appendix~\ref{app:baseline-validation}; component ablations in Section~\ref{sec:val-ablation}.

\subsection{Main results}
\label{sec:main-bench}

Table~\ref{tab:main-bench} reports external benchmark accuracy together with
unique-prompt exposure. On ViRL39K, averaged over three seeds, \method{} reaches a $13.25$ four-benchmark
average, a $\mathbf{+32.4\%}$ relative improvement over Random-$10\%$ ($10.01$)
and a $\mathbf{+68.6\%}$ improvement over the Qwen3-VL-2B base model ($7.86$).
On VLAA-Thinking, \method{} reaches $13.19$, $\mathbf{+16.5\%}$ above the
corresponding Random-$10\%$ ($11.32$) and all other baselines. Per-seed results for \method{} and Random at the 5\% and 10\% budgets are in Appendix~\ref{app:seeds}, where \method{}-$10\%$ outperforms Random-$10\%$ in every run. A paired bootstrap over matched evaluation items gives strictly positive 95\% intervals for \method{}-10\% against Random-10\%, MaxVar-10\%, Hard-10\%, and full-data GRPO (Appendix~\ref{app:benchmark-bootstrap}), complementing the training-seed results. Across seeds, Random-5\% and Random-10\% are tied (9.96 vs.\ 10.01), while \method{}-5\% is noisier and slightly below \method{}-10\% (12.93 vs.\ 13.25). With a fixed step budget, the 5\% pool is revisited about twice as often and freezes midway through training, so the policy fits it more tightly and results depend more on which prompts were selected. Gains are also not uniform across benchmarks. For example, on VLAA-Thinking MATH-Vision, \method{}-10\% (4.93) trails full-data GRPO (6.25). Unlike Hard-$\beta$ and MaxVar-$\beta$, which train on 10\% of prompts but score the entire pool, \method{} avoids full-pool
verifier scans and reaches the best overall average with only $1.16\%$
Selector-first exposure on ViRL39K (observed Probe-only exposure $0.43\%$). Full-data GRPO trails the best budgeted selector under the
matched step budget. It spreads updates over many prompts, including all-incorrect groups that carry zero group-relative advantage, whereas \method{} concentrates updates on mixed-reward prompts. This reflects signal-efficient selection under fixed compute rather than a general benefit of smaller datasets.

\begin{table}[t]
\centering
\caption{External benchmark accuracy (\%) with verified-prompt exposure and selector cost. Exposure is a percentage of the full source pool ($N$ prompts), split by acquisition (\textbf{Selector-first}, \textbf{Training-first}) and by final use (\textbf{Train}, \textbf{Probe-only});  both views sum to \textbf{Total} (Appendix~\ref{app:prompt-exposure}). Range cells give deterministic admission bounds (not confidence intervals), with the observed value (obs.) below. Dense-scored baselines scan the 90\% training split and are not exposure-matched to strict-budget methods. \textbf{Probe Gen.}\ counts selector-only generations as a multiple of $N$. Accuracies use a 512-token generation cap. Bold marks the best result per dataset.}
\label{tab:main-bench}
\scriptsize
\setlength{\tabcolsep}{1.6pt}
\renewcommand{\arraystretch}{1.12}
\begin{adjustbox}{max width=\textwidth}
\begin{tabular}{>{\centering\arraybackslash}m{0.72cm} l!{\vrule width 0.25pt} c c c c c !{\vrule width 0.25pt} c !{\vrule width 0.25pt} c c c c c}
\toprule
\textbf{Dataset} & \textbf{Method}
& \multicolumn{5}{!{\vrule width 0.25pt}c!{\vrule width 0.25pt}}{\textbf{Verified-prompt exposure (\%) $\downarrow$}}
& \multicolumn{1}{c!{\vrule width 0.25pt}}{\textbf{Selector compute $\downarrow$}}
& \multicolumn{5}{c}{\textbf{External benchmark accuracy (\%) $\uparrow$}} \\
\cmidrule(lr){3-7}\cmidrule(lr){8-8}\cmidrule(lr){9-13}
& & \multicolumn{2}{c}{\textbf{By acquisition}} & \multicolumn{2}{c}{\textbf{By final use}} & & & & & & & \\
\cmidrule(lr){3-4}\cmidrule(lr){5-6}
& & \textbf{Selector-} & \textbf{Training-} & \textbf{Train} & \textbf{Probe-} & \textbf{Total}
& \textbf{Probe Gen.}
& \textbf{WeMath} & \textbf{MathVerse} & \textbf{MATH-Vision} & \textbf{DynaMath} & \textbf{Avg.} \\
& & \textbf{first} & \textbf{first} & & \textbf{only} &
& \textbf{($\times N$)}
& & & & & \\
\midrule
Base & Qwen3-VL-2B
& -- & -- & -- & -- & -- & --
& 1.78 & 8.81 & 4.61 & 16.23 & 7.86 \\
\midrule
\multirow{10}{*}{\rotatebox[origin=c]{90}{ViRL39K \citep{wang2025virl}}}
& Random-$5\%$
& 0.00 & 5.00 & 5.00 & 0.00 & 5.00 & 0
& 11.26 & 10.18 & 5.61 & 16.72 & 10.94 \\
& Random-$10\%$
& 0.00 & 10.00 & 10.00 & 0.00 & 10.00 & 0
& 10.12 & 9.66 & 5.59 & 16.14 & 10.38 \\
& Hard-$10\%$
& 90.00 & 0.00 & 10.00 & 80.00 & 90.00 & 0.90
& 1.67 & 9.75 & 7.24 & 16.31 & 8.74 \\
& MaxVar-$10\%$
& 90.00 & 0.00 & 10.00 & 80.00 & 90.00 & 7.20
& 6.49 & 10.71 & \textbf{9.87} & 17.09 & 11.04 \\
& DOTS-MM-$10\%$ \citep{sun2025dots}
& 3.36 & 6.64 & \rng{6.64,10.00}{9.68} & \rng{0.00,3.36}{0.32} & 10.00 & 0.27
& 3.39 & 8.93 & 7.24 & 16.93 & 9.12 \\
& VADE original \citep{hu2025vade}
& 90.00 & 0.00 & 47.77 & 42.23 & 90.00 & 7.20
& 0.92 & 9.59 & 6.25 & 16.51 & 8.32 \\
& VADE-$10\%$ \citep{hu2025vade}
& 90.00 & 0.00 & 10.00 & 80.00 & 90.00 & 7.20
& 5.86 & 9.16 & 6.25 & 15.35 & 9.15 \\
& Full train split
& 0.00 & 90.00 & 90.00 & 0.00 & 90.00 & 0
& 6.32 & 9.92 & 5.92 & 16.67 & 9.71 \\
\cmidrule(lr){2-13}
& \method{}-$5\%$
& 1.16 & 3.84 & \rng{3.84,5.00}{4.49} & \rng{0.00,1.16}{0.51} & 5.00 & 0.09
& 6.85 & 10.53 & 7.56 & 17.21 & 10.53 \\
& \textbf{\method{}-$10\%$}
& 1.16 & 8.84 & \rng{8.84,10.00}{9.57} & \rng{0.00,1.16}{0.43} & 10.00 & 0.09
& \textbf{13.45} & \textbf{11.12} & 7.57 & \textbf{20.32} & \textbf{13.12} \\
\midrule
\multirow{10}{*}{\rotatebox[origin=c]{90}{VLAA-Thinking \citep{chen2504vlaa}}}
& Random-$5\%$
& 0.00 & 5.00 & 5.00 & 0.00 & 5.00 & 0
& 11.32 & 12.16 & 5.92 & 20.10 & 12.38 \\
& Random-$10\%$
& 0.00 & 10.00 & 10.00 & 0.00 & 10.00 & 0
& 8.68 & 11.57 & 5.26 & 19.76 & 11.32 \\
& Hard-$10\%$
& 90.00 & 0.00 & 10.00 & 80.00 & 90.00 & 0.90
& 4.25 & 11.47 & 5.02 & 18.74 & 9.87 \\
& MaxVar-$10\%$
& 90.00 & 0.00 & 10.00 & 80.00 & 90.00 & 7.20
& 7.83 & 12.05 & 6.19 & 20.96 & 11.76 \\
& DOTS-MM-$10\%$
& 4.60 & 5.40 & \rng{5.40,10.00}{9.56} & \rng{0.00,4.60}{0.44} & 10.00 & 0.37
& 4.10 & 10.76 & 5.72 & 20.38 & 10.24 \\
& VADE original
& 90.00 & 0.00 & 47.25 & 42.75 & 90.00 & 7.20
& 1.62 & 11.18 & 4.84 & 20.08 & 9.43 \\
& VADE-$10\%$
& 90.00 & 0.00 & 10.00 & 80.00 & 90.00 & 7.20
& 6.52 & 10.90 & 4.80 & 18.92 & 10.29 \\
& Full train split
& 0.00 & 90.00 & 90.00 & 0.00 & 90.00 & 0
& 9.25 & 11.73 & \textbf{6.25} & 21.30 & 12.13 \\
\cmidrule(lr){2-13}
& \method{}-$5\%$
& 1.59 & 3.41 & \rng{3.41,5.00}{4.74} & \rng{0.00,1.59}{0.26} & 5.00 & 0.13
& 8.24 & 12.52 & 6.04 & 20.11 & 11.73 \\
& \textbf{\method{}-$10\%$}
& 1.59 & 8.41 & \rng{8.41,10.00}{9.80} & \rng{0.00,1.59}{0.20} & 10.00 & 0.13
& \textbf{13.05} & \textbf{13.45} & 4.93 & \textbf{21.34} & \textbf{13.19} \\
\bottomrule
\end{tabular}
\end{adjustbox}
\end{table}

\subsection{Component ablation}
\label{sec:val-ablation}

\begin{table}[t]
\centering
\caption{Component ablations on ViRL39K under the 10\% GRPO training-prompt
cap. Mechanism columns indicate which selector modules are enabled.
\textbf{Avg.}\ is the four-benchmark external average, and \textbf{Ent.}\ is
final actor-policy entropy. Cadence denotes how often new prompts enter the training pool: default is the main adaptive schedule, slow refreshes more conservatively, and one-shot stops after the initial discovery pass.
Each row reports one training run.
}
\label{tab:components}
\scriptsize
\setlength{\tabcolsep}{2.8pt}
\renewcommand{\arraystretch}{1.13}
\begin{adjustbox}{max width=\textwidth}
\begin{tabular}{c l l cccccc}
\toprule
\textbf{Group} & \textbf{Intervention} & \textbf{Tests} & \textbf{Transfer} & \textbf{Utility} & \textbf{Refresh} & \textbf{Cadence} & \textbf{Avg. $\uparrow$} & \textbf{Ent.} \\
\midrule
\multirow{1}{*}{Reference}
& Full \method{}
& Reference setting
& KNN & asym & \checkmark & default
&  $\mathbf{13.12}$ & $0.20$ \\
\midrule
\multirow{7}{*}{Predictor}
& Coverage only
& Coverage without utility
& \xmark & \xmark & \xmark & one-shot
&  $8.23$ & $0.40$ \\
& One-shot KNN
& One-shot vs. online KNN
& KNN & asym & \xmark & one-shot
&  $10.03$ & $0.30$ \\
& MLP transfer
& Learned vs. local transfer
& MLP & asym & \checkmark & default
&  $11.32$ & $0.23$ \\
& Ridge transfer
& Linear vs. local transfer
& Ridge & asym & \checkmark & default
&  $8.80$ & $0.28$ \\
& Ridge active probes
& More probes for Ridge
& Ridge+AP & asym & \checkmark & default
&  $10.86$ & $0.09$ \\
& Random score
& Random predictor
& random & asym & \checkmark & default
& $10.02$ & $0.26$ \\
& Random score (slow)
& Random predictor
& random & asym & \checkmark & slow
& $10.76$ & $0.36$ \\
\midrule
\multirow{2}{*}{Utility}
& No hard-side bias
& Asymmetric utility bias
& KNN & symmetric & \checkmark & default
&  $10.43$ & $0.38$ \\
& Dead-zone $h{=}0.8$
& Easy-filter aggressiveness
& KNN & $h{=}0.8$ & \checkmark & default
&  $11.12$ & $0.12$ \\
\midrule
\multirow{1}{*}{Cadence}
% \multirow{2}{*}{Cadence}
% & Aggressive cadence
% & Rapid-admission stress test
% & KNN & asym & \checkmark & fast
% &  $9.09$  & $0.09$ \\
& Slow cadence
& Delayed-freeze control
& KNN & asym & \checkmark & slow
&  $9.51$  & $0.28$ \\
% \midrule
% \multirow{2}{*}{Visual}
% & Default $+$ \vll{}
% & Visual gate on default
% & KNN & asym & \checkmark & \checkmark & default
% &  $9.67$ & $0.31$ \\
% & Slow $+$ \vll{}
% & Visual gate on slow
% & KNN & asym & \checkmark & \checkmark & slow
% &  $9.22$ & $0.22$ \\
\bottomrule
\end{tabular}
\end{adjustbox}
\end{table}

Table~\ref{tab:components} ablates ViRL39K components under the same 10\% budget, supporting the following claims.

\paragraph{(a) Geometric coverage is necessary but insufficient.}
Cluster-only selection falls below the Random-$10\%$ baseline, indicating that spreading the budget across clusters is ineffective without a per-prompt utility signal. On curated multimodal corpora, many in-distribution prompts are already saturated under the base policy, and coverage alone cannot filter them out.

\paragraph{(b) Sparse-to-dense transfer makes the budget effective.}
All transfer controls fall below the full method. Under the default cadence, random scoring ($10.02$), Ridge, Ridge
with active probes, and MLP transfer each lose accuracy despite using the same
budgeted discovery loop. The gain does not come from coverage or
online refresh alone, but from transferring sparse rollout feedback through
local embedding neighborhoods. Even the learned MLP predictor underperforms
cosine-KNN, consistent with diagnostics in Section~\ref{sec:predictor-diag}:
Frozen multimodal embeddings provide useful ranking signals, while a parametric head is harder to fit on a small, evolving online buffer.

\paragraph{(c) The pool must refresh as the policy evolves.}
Immediate freeze runs the KNN ranker only at initialization and then stops admitting new prompts, losing over three points relative to our proposed method. This suggests the round-$0$ KNN signal does not reliably predict which prompts will be useful later and closed-loop refresh is needed to keep utility estimates aligned with the current policy.

\paragraph{(d) Discovery cadence affects selector quality.}
Cadence controls how often the selector admits new training prompts. The
headline schedule refreshes the pool frequently early in training, then slows
admission as the budget fills. The slow control is more conservative, admitting
fewer new prompts with longer gaps between refreshes. Its lower performance with the same KNN transfer and utility shows that early pool
growth matters. Delaying admission keeps the policy on a narrow pool for too long
and starves the ranker of fresh feedback. Cadence also interacts with the predictor. Moving from the slow to the default schedule raises KNN from $9.51$ to $13.12$ but lowers random scoring from $10.76$ to $10.02$, so KNN transfer benefits most when its estimates are refreshed frequently.

\paragraph{(e) Utility shape contributes a smaller but consistent signal.}
Removing the asymmetric hard-side bias raises final entropy to $0.38$
without late-training collapse, indicating selection has become more
diffuse; tightening the dead-zone with $h{=}0.8$ collapses entropy to
$0.12$, indicating over-concentration on a narrow band. Both deviations
from the default $[0.05,0.95]$ band degrade the trajectory.

% \paragraph{(f) The visual-dependence gate is a diagnostic, not the
% source of the gain.}
% Adding \vll{} on top of the main configuration drops bench avg under
% both cadences ($9.67$ fast, $9.22$ slow). This is consistent with our
% audit (Section~\ref{sec:audit}): on ViRL39K, after cluster-balanced
% discovery has already been applied, the residual fraction of text-leaky
% prompts is small, so an extra multiplicative gate suppresses
% borderline-but-useful prompts more than it suppresses true text-prior
% shortcuts. We therefore retain \vll{} only as an analysis channel here;
% its role is more clearly motivated for visually heterogeneous corpora,
% where the audit shows a much larger text-leaky tail.
\subsection{Visual dependence and reward variance}
\label{sec:audit}
\vspace{-0.1in}
As a diagnostic, we test whether visual dependence is already captured by reward variance. We audit a random subset of ViRL39K
prompts ($n{=}2{,}000$) by running $G{=}8$ text-only rollouts per prompt with
the same Qwen3-VL-2B base model. For each prompt, we compare the with-image
reward mean $R_i^{\mathrm{img}}$ with the text-only reward mean
$R_i^{\varnothing}$, record the with-image reward variance
$v_i^{\mathrm{img}}$, and define the audit reward gain
$\Delta R_i=R_i^{\mathrm{img}}-R_i^{\varnothing}$.

The audit shows that $24.2\%$ of prompts are solved from text alone, while only
$31.8\%$ are clearly helped by the image. Moreover, the audit did not detect a monotonic association between text-only
reward and with-image reward variance (Spearman $\rho={-}0.020$, $p{=}0.37$).
This does not establish independence. Visual dependence is also geometrically structured, as
cluster identity explains more visual lift than reward variance
($\eta^2{=}0.282$ vs. $0.113$). These results motivate estimating visual
dependence separately from reward variance.

\paragraph{Held-out audit probe.}
We compare \method{}-10\% with the Base model before RL on a held-out ViRL39K audit fold, using eight rollouts per prompt and the training verifier. Accuracy rises from $41.4\%$ to $58.3\%$ on all audited prompts ($+16.9$ pp, 95\% paired-bootstrap CI $[13.8, 19.9]$) and from $63.0\%$ to $79.5\%$ on image-dependent prompts ($+16.6$ pp, $[11.4, 21.7]$). Gains are similar on non-visual/mixed ($+17.1$ pp, $[13.3, 20.9]$) and text-leaky ($+17.8$ pp, $[11.2, 24.7]$) prompts, and the relative error reduction is larger on image-dependent prompts ($44.8\%$ vs.\ $28.8\%$ overall). These in-distribution accuracies use the training verifier and are not directly comparable to external benchmarks in Table~\ref{tab:main-bench}, but show that \method{} improves visually grounded prompts, not just text-solvable ones.

\begin{figure}[t]
\centering
\includegraphics[width=\linewidth]{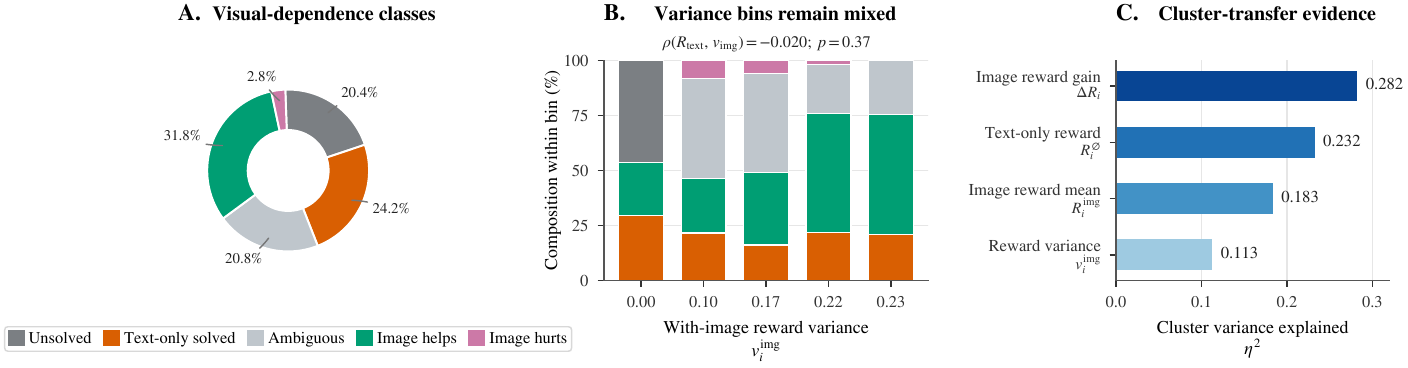}
\vspace{-3mm}
\caption{
Visual-dependence audit on ViRL39K prompts. (A) Mutually exclusive audit classes; $24.2\%$ are solvable from text alone. (B) Class composition across with-image reward variance $v_i^{\mathrm{img}}$ bins. No monotonic association was detected between text-only reward and with-image reward variance ($\rho={-}0.020$, $p{=}0.37$). (C) Cluster-level explained variance ($\eta^2$): embedding clusters explain more variance in image reward gain $\Delta R_i$ than in $v_i^{\mathrm{img}}$ ($0.282$ vs. $0.113$).\vspace{-0.2in}}
\label{fig:visual-dependence}
\end{figure}

\subsection{Selector diagnostics}
\label{sec:predictor-diag}
% \vspace{-0.1in}
Finally, selector diagnostics assess whether the online loop behaves as intended. Predictor quality is measured with leave-one-out rank
correlation rather than training $R^2$, which measures fit to observed data. In
the slow-cadence run, leave-one-out Spearman reaches $0.21$ for disagreement and
$0.39$ for the mean by step $250$. Table~\ref{tab:selector-transfer}
summarizes leave-one-out correlations across selector variants.

We also examine whether the frozen embeddings capture difficulty and support
utility transfer beyond the selector buffer. Clustering reduces within-cluster
difficulty variation by $7.8\%$ relative to 500 size-preserving shuffled
controls ($p<0.002$). However, $76.8\%$ of evaluated clusters include both fully
solved and fully failed prompts, so cluster membership alone does not determine
difficulty.

On a held-out ViRL39K set excluded from training and selector observations,
neighbor-predicted utility correlates with measured utility: Spearman rises from
$0.163$ for the base model to $0.364$ for \method{} with unchanged embeddings,
while random neighbors give correlations near zero. \method{}'s highest-ranked
$10\%$ have $48\%$ higher realized utility than the held-out-set average (95\%
bootstrap interval $[17\%, 80\%]$). Appendix~\ref{app:embedding-clusters}
reports confidence intervals, neighborhood-size and representation checks, and
uniform-outcome prediction results (analyses limited to Qwen3-VL-2B).
\section{Discussion}
\vspace{-0.1in}
\label{sec:discussion}

\noindent\textbf{Interpreting the ablations.}
The ablations show that budgeted GRPO requires controlling three complementary
failure modes. Cluster coverage prevents the budget from collapsing onto a
narrow semantic region; online refresh prevents KNN transfer from relying on a
stale round-$0$ policy snapshot; and stable cadence prevents the adaptive ranker
from overreacting to outdated rollout-buffer targets. Together, coverage
determines where the pool can grow, KNN estimates which prompts are currently
useful, and cadence keeps those estimates synchronized with the policy.

% \paragraph{Interpreting the ablations.}
% The results should not be read as a claim that one scorer alone solves
% budgeted GRPO. They show that the selector has to control three distinct
% failure modes. Cluster coverage prevents the budget from collapsing onto
% a narrow semantic region. Online refresh prevents KNN transfer from
% being anchored to a small round-$0$ snapshot of the policy. Stable
% cadence prevents an adaptive ranker from overreacting to a rollout
% buffer whose targets have become stale. These mechanisms are
% complementary: coverage chooses where the pool can grow, KNN estimates
% which prompts are currently useful, and cadence determines whether those
% estimates remain synchronized with the policy.

\noindent\textbf{Implications for multimodal selection.}
Static data pruning treats prompt value as fixed, but GRPO changes the
value of a prompt as the model learns. This makes budgeted selection a
closed-loop control problem rather than a one-shot subset problem.
Multimodal data adds a second distinction: a prompt can be useful for
GRPO because it induces reward disagreement, yet still be solvable from
text priors. The text-only audit did not detect a monotonic association between text-only reward and with-image reward variance, highlighting the need for visual-dependence checks alongside benchmark gains.
% Future work could
% revisit VLL-like image-sensitivity signals as a regularized auxiliary channel;
% we treat them here as diagnostics rather than a source of headline gains.

\noindent\textbf{Scope of transfer.}
\method{} is most useful when samples are heterogeneous, verifiable feedback is
expensive, and frozen multimodal embeddings provide meaningful neighborhoods.
It is less beneficial when the pool is small enough for exhaustive training, is
highly homogeneous, or has embedding neighborhoods that poorly match capability
transfer. The text-only audit is corpus-agnostic and can be applied to other
multimodal RL datasets to decide whether a visual-dependence channel is needed.
% \noindent\textbf{Scope of transfer.}
% \method{} is most appropriate when prompts are heterogeneous, verifiable
% feedback is expensive, and frozen multimodal embeddings preserve useful
% neighborhood structure. It may be less valuable when the pool is small
% enough to train on exhaustively, when the dataset is already highly
% homogeneous, or when embedding neighborhoods do not align with
% capability transfer. The audit methodology itself is corpus-agnostic:
% other multimodal RL datasets can be checked with text-only rollouts
% before deciding whether a visual-dependence channel is needed.

\noindent\textbf{Limitations and future directions.}
Our evaluation focuses on a single model variant, Qwen3-VL-2B;
extending the study to additional model families and scales
remains future work. On ViRL39K, the advantage over
budget-matched random selection at the 10\% budget is
consistent across three runs, although three seeds do not fully
characterize GRPO, initialization, or clustering variability. The greater run-to-run
variability at the 5\% budget suggests that establishing
robustness under tighter budget constraints would require
additional evaluation across training seeds. VLAA-Thinking
experiments and component ablations use one run per
configuration; further replication would help characterize
their variability.

\method{} targets heterogeneous prompts with costly verifier feedback and frozen multimodal embeddings that support neighborhood transfer. It requires lightweight preparation: cached embeddings, semantic clusters, and representative probes to initialize the online buffer. Current implementation uses fixed, empirically validated hyperparameters and choices. Future work could adapt these to evolving policies and jointly optimize verified-prompt exposure, rollout cost, semantic coverage, grounding diagnostics, and benchmark-specific transfer within a multi-objective allocation framework.
\vspace{-0.1in}
\section{Conclusion}
\vspace{-0.1in}
\label{sec:conclusion}

% We introduced \method, an online data-discovery framework for budgeted
% multimodal GRPO. Instead of treating the training pool as fixed,
% \method{} spends a limited unique-prompt budget on examples that are
% predicted to be useful for the current policy: frozen multimodal
% geometry preserves coverage, recency-weighted rollouts estimate
% policy-dependent reward statistics, and cosine-KNN transfers sparse feedback to unobserved prompts. The resulting discovery-to-freeze
% controller builds a compact training pool and then continues to reweight
% it as the policy changes.

% On ViRL39K, \textsc{Discover}-10\% outperforms Random-10\% in all three training runs and, for fixed checkpoints, has positive paired-bootstrap intervals against Random-10\%, the dense-scored MaxVar-10\% and Hard-10\%, and full-data GRPO. Ablations show that success requires integrating coverage, online updates, and cadence, as no single ranker is sufficient. Coverage, online refresh, and cadence address different failure modes. The text-only audit did not detect a monotonic association between text-only reward and with-image reward variance. These findings suggest treating multimodal RL post-training as dynamic budget allocation under coverage and grounding constraints rather than one-shot data pruning.

We introduced \method{}, an online data-discovery framework for budgeted multimodal GRPO that selects prompts useful to the current policy under a limited unique-prompt budget. Frozen multimodal geometry maintains coverage, recency-weighted rollouts estimate policy-dependent reward statistics, and cosine-KNN transfers sparse feedback to unobserved prompts. A discovery-to-freeze controller builds a compact pool, reweighting it as the policy evolves.

On ViRL39K, \textsc{Discover}-10\% outperforms Random-10\% across all three runs, with positive paired-bootstrap intervals at fixed checkpoints against Random-10\%, dense-scored MaxVar-10\% and Hard-10\%, and full-data GRPO. Ablations show that coverage, online updates, and cadence address distinct failure modes, with no single ranker sufficient. The text-only audit did not detect a monotonic association between text-only reward and with-image reward variance. These findings frame RL post-training as dynamic budget allocation under coverage and grounding constraints, rather than one-shot data pruning.

\begin{ack}
This work was funded, in part, by the Vector Institute for AI, Canada CIFAR AI Chair, NSERC Canada Research Chair (CRC), and NSERC Discovery Grant RGPIN-2019-05624. Resources used in preparing this research were provided, in part, by the Province of Ontario, the Government of Canada through CIFAR, the Digital Research Alliance of Canada, companies\footnote{\url{https://vectorinstitute.ai/\#partners}} sponsoring the Vector Institute, and Advanced Research Computing at the University of British Columbia. Additional hardware support was provided by John R. Evans Leaders Fund CFI grant and Digital Research Alliance of Canada under the Resource Allocation Competition award, as well as NIH grant 1R01NS138765-01 and the Ovarian Cancer Research Alliance (OCRA) grant CRDGAI-2023-2-2101. The authors declare no competing interests.
\end{ack}

\bibliographystyle{plainnat}
\bibliography{references}

@article{schulman2017ppo,
  title = {Proximal Policy Optimization Algorithms},
  author = {Schulman, John and Wolski, Filip and Dhariwal, Prafulla and Radford, Alec and Klimov, Oleg},
  journal = {arXiv preprint arXiv:1707.06347},
  year = {2017},
}

@article{shao2024deepseekmath,
  title = {{DeepSeekMath}: Pushing the Limits of Mathematical Reasoning in Open Language Models},
  author = {Shao, Zhihong and Wang, Peiyi and Zhu, Qihao and Xu, Runxin and Song, Junxiao and Bi, Xiao and Zhang, Haowei and Zhang, Mingchuan and Li, Y. K. and Wu, Y. and Guo, Daya},
  journal = {arXiv preprint arXiv:2402.03300},
  year = {2024},
}

@inproceedings{sun2025dots,
  title = {Improving Data Efficiency for {LLM} Reinforcement Fine-tuning Through Difficulty-targeted Online Data Selection and Rollout Replay},
  author = {Sun, Yifan and Shen, Jingyan and Wang, Yibin and Chen, Tianyu and Wang, Zhendong and Zhou, Mingyuan and Zhang, Huan},
  booktitle = {Advances in Neural Information Processing Systems},
  volume = {38},
  year = {2025}
}

@inproceedings{zheng2025greso,
  title = {Act Only When It Pays: Efficient Reinforcement Learning for {LLM} Reasoning via Selective Rollouts},
  author = {Zheng, Haizhong and Zhou, Yang and Bartoldson, Brian R. and Kailkhura, Bhavya and Lai, Fan and Zhao, Jiawei and Chen, Beidi},
  booktitle = {Advances in Neural Information Processing Systems},
  volume = {38},
  year = {2025}
}

@inproceedings{evans2024jest,
  title = {Data Curation via Joint Example Selection Further Accelerates Multimodal Learning},
  author = {Evans, Talfan and Parthasarathy, Nikhil and Merzi{\'c}, Hamza and H{\'e}naff, Olivier J.},
  booktitle = {Advances in Neural Information Processing Systems},
  volume = {37},
  year = {2024},
}

@inproceedings{xia2024less,
  title = {{LESS}: Selecting Influential Data for Targeted Instruction Tuning},
  author = {Xia, Mengzhou and Malladi, Sadhika and Gururangan, Suchin and Arora, Sanjeev and Chen, Danqi},
  booktitle = {Proceedings of the 41st International Conference on Machine Learning},
  pages = {54104--54132},
  year = {2024},
  volume = {235},
  series = {Proceedings of Machine Learning Research},
  publisher = {PMLR},
}

@inproceedings{yang2024s2l,
  title = {{SmallToLarge} ({S2L}): Scalable Data Selection for Fine-tuning Large Language Models by Summarizing Training Loss Trajectories of Small Models},
  author = {Yang, Yu and Mishra, Siddhartha and Chiang, Jeffrey and Mirzasoleiman, Baharan},
  booktitle = {Advances in Neural Information Processing Systems},
  volume = {37},
  year = {2024},
}

@inproceedings{liu2024tsds,
  title = {{TSDS}: Data Selection for Task-Specific Model Finetuning},
  author = {Liu, Zifan and Karbasi, Amin and Rekatsinas, Theodoros},
  booktitle = {Advances in Neural Information Processing Systems},
  volume = {37},
  year = {2024},
}

@inproceedings{zhang2024tagcos,
  title = {{TAGCOS}: Task-agnostic Gradient Clustered Coreset Selection for Instruction Tuning Data},
  author = {Zhang, Jipeng and Qin, Yaxuan and Pi, Renjie and Zhang, Weizhong and Pan, Rui and Zhang, Tong},
  booktitle = {Findings of the Association for Computational Linguistics: NAACL 2025},
  pages = {4686--4701},
  year = {2025}
}

@article{liu2024tive,
  title = {Less is More: High-value Data Selection for Visual Instruction Tuning},
  author = {Liu, Zikang and Zhou, Kun and Zhao, Wayne Xin and Gao, Dawei and Li, Yaliang and Wen, Ji-Rong},
  journal = {arXiv preprint arXiv:2403.09559},
  year = {2024},
}

@inproceedings{bengio2009curriculum,
  title = {Curriculum Learning},
  author = {Bengio, Yoshua and Louradour, J{\'e}r{\^o}me and Collobert, Ronan and Weston, Jason},
  booktitle = {Proceedings of the 26th International Conference on Machine Learning},
  pages = {41--48},
  year = {2009}
}

@inproceedings{kumar2010selfpaced,
  title = {Self-Paced Learning for Latent Variable Models},
  author = {Kumar, M. Pawan and Packer, Benjamin and Koller, Daphne},
  booktitle = {Advances in Neural Information Processing Systems},
  volume = {23},
  year = {2010}
}

@inproceedings{saran2023vessal,
  title = {Streaming Active Learning with Deep Neural Networks},
  author = {Saran, Akanksha and Yousefi, Safoora and Krishnamurthy, Akshay and Langford, John and Ash, Jordan T.},
  booktitle = {Proceedings of the 40th International Conference on Machine Learning},
  pages = {30005--30021},
  year = {2023},
  volume = {202},
  series = {Proceedings of Machine Learning Research},
  publisher = {PMLR},
}

@inproceedings{wang2025virl,
  title = {{VL-Rethinker}: Incentivizing Self-Reflection of Vision-Language Models with Reinforcement Learning},
  author = {Wang, Haozhe and Qu, Chao and Huang, Zuming and Chu, Wei and Lin, Fangzhen and Chen, Wenhu},
  booktitle = {Advances in Neural Information Processing Systems},
  volume = {38},
  year = {2025}
}

@inproceedings{zhang2025r1vl,
  title = {{R1-VL}: Learning to Reason with Multimodal Large Language Models via Step-wise Group Relative Policy Optimization},
  author = {Zhang, Jingyi and Huang, Jiaxing and Yao, Huanjin and Liu, Shunyu and Zhang, Xikun and Lu, Shijian and Tao, Dacheng},
  booktitle = {Proceedings of the IEEE/CVF International Conference on Computer Vision},
  year = {2025}
}

@article{chen2504vlaa,
  title={{SFT} or {RL}? An Early Investigation into Training {R1}-Like Reasoning Large Vision-Language Models},
  author={Chen, Hardy and Tu, Haoqin and Wang, Fali and Liu, Hui and Tang, Xianfeng and Du, Xinya and Zhou, Yuyin and Xie, Cihang},
  journal={arXiv preprint arXiv:2504.11468},
  year={2025}
}

@inproceedings{huang2025visionr1,
  title = {{Vision-R1}: Incentivizing Reasoning Capability in Multimodal Large Language Models},
  author = {Huang, Wenxuan and Jia, Bohan and Zhai, Zijie and Cao, Shaosheng and Ye, Zheyu and Zhao, Fei and Xu, Zhe and Tang, Xu and Hu, Yao and Lin, Shaohui},
  booktitle = {International Conference on Learning Representations},
  year = {2026}
}

@inproceedings{zhang2024mathverse,
  title = {{MathVerse}: Does Your Multi-modal {LLM} Truly See the Diagrams in Visual Math Problems?},
  author = {Zhang, Renrui and Jiang, Dongzhi and Zhang, Yichi and Lin, Haokun and Guo, Ziyu and Qiu, Pengshuo and Zhou, Aojun and Lu, Pan and Chang, Kai-Wei and Gao, Peng and Li, Hongsheng},
  booktitle = {European Conference on Computer Vision},
  year = {2024}
}

@inproceedings{wang2024mathvision,
  title = {Measuring Multimodal Mathematical Reasoning with {MATH-Vision} Dataset},
  author = {Wang, Ke and Pan, Junting and Shi, Weikang and Lu, Zimu and Ren, Houxing and Zhou, Aojun and Zhan, Mingjie and Li, Hongsheng},
  booktitle = {Advances in Neural Information Processing Systems},
  volume = {37},
  year = {2024},
}

@inproceedings{qiao2025wemath,
  title = {{We-Math}: Does Your Large Multimodal Model Achieve Human-like Mathematical Reasoning?},
  author = {Qiao, Runqi and Tan, Qiuna and Dong, Guanting and Wu, Minhui and Sun, Chong and Song, Xiaoshuai and Wang, Jiapeng and GongQue, Zhuoma and Lei, Shanglin and Zhang, YiFan and Wei, Zhe and Zhang, Miaoxuan and Qiao, Runfeng and Zong, Xiao and Xu, Yida and Yang, Peiqing and Bao, Zhimin and Diao, Muxi and Li, Chen and Zhang, Honggang},
  booktitle = {Proceedings of the 63rd Annual Meeting of the Association for Computational Linguistics (Volume 1: Long Papers)},
  pages = {20023--20070},
  year = {2025},
  address = {Vienna, Austria},
  publisher = {Association for Computational Linguistics},
}

@inproceedings{zou2025dynamath,
  title = {{DynaMath}: A Dynamic Visual Benchmark for Evaluating Mathematical Reasoning Robustness of Vision Language Models},
  author = {Zou, Chengke and Guo, Xingang and Yang, Rui and Zhang, Junyu and Hu, Bin and Zhang, Huan},
  booktitle = {International Conference on Learning Representations},
  year = {2025},
}

@article{bai2025qwen3vl,
  title = {{Qwen3-VL} Technical Report},
  author = {Bai, Shuai and Cai, Yuxuan and others},
  journal = {arXiv preprint arXiv:2511.21631},
  year = {2025},
}

@inproceedings{yu2025dapo,
  title={{DAPO}: An Open-Source {LLM} Reinforcement Learning System at Scale},
  author={Yu, Qiying and Zhang, Zheng and others},
  booktitle={Advances in Neural Information Processing Systems},
  volume={38},
  year={2025}
}

@article{li2026qwen3vlembedding,
  title={{Qwen3-VL-Embedding} and {Qwen3-VL-Reranker}: A Unified Framework for State-of-the-Art Multimodal Retrieval and Ranking},
  author={Li, Mingxin and Zhang, Yanzhao and Long, Dingkun and Chen, Keqin and Song, Sibo and Bai, Shuai and Yang, Zhibo and Xie, Pengjun and Yang, An and Liu, Dayiheng and Zhou, Jingren and Lin, Junyang},
  journal={arXiv preprint arXiv:2601.04720},
  year={2026}
}

@techreport{settles2009active,
  title={Active Learning Literature Survey},
  author={Settles, Burr},
  institution={University of Wisconsin--Madison Department of Computer Sciences},
  number={1648},
  type={Computer Sciences Technical Report},
  year={2009}
}

@inproceedings{sener2018coreset,
  title={Active Learning for Convolutional Neural Networks: A Core-Set Approach},
  author={Sener, Ozan and Savarese, Silvio},
  booktitle={International Conference on Learning Representations},
  year={2018}
}

@article{guo2025deepseekr1,
  title={{DeepSeek-R1} Incentivizes Reasoning in {LLMs} through Reinforcement Learning},
  author={{DeepSeek-AI} and Guo, Daya and Yang, Dejian and Zhang, Haowei and Song, Junxiao and others},
  journal={Nature},
  volume={645},
  pages={633--638},
  year={2025}
}

@inproceedings{liu2025drgrpo,
  title={Understanding {R1-Zero}-Like Training: A Critical Perspective},
  author={Liu, Zichen and Chen, Changyu and Li, Wenjun and Qi, Penghui and Pang, Tianyu and Du, Chao and Lee, Wee Sun and Lin, Min},
  booktitle={Conference on Language Modeling},
  year={2025}
}

@article{baroian2026promptreplay,
  title={Prompt Replay: Speeding up {GRPO} with On-Policy Reuse of High-Signal Prompts},
  author={Baroian, Andrei and Berger, Rutger},
  journal={arXiv preprint arXiv:2603.21177},
  year={2026}
}

@article{li2025limr,
  title={{LIMR}: Less is More for {RL} Scaling},
  author={Li, Xuefeng and Zou, Haoyang and Liu, Pengfei},
  journal={arXiv preprint arXiv:2502.11886},
  year={2025}
}

@inproceedings{duan2024vlmevalkit,
  title={{VLMEvalKit}: An Open-Source Toolkit for Evaluating Large Multi-Modality Models},
  author={Duan, Haodong and Yang, Junming and Qiao, Yuxuan and Fang, Xinyu and Chen, Lin and Liu, Yuan and Agarwal, Amit and Chen, Zhe and Li, Mo and Ma, Yubo and others},
  booktitle={Proceedings of the 32nd ACM International Conference on Multimedia},
  pages={11198--11201},
  year={2024}
}

@inproceedings{gao2025prompt,
  title={Prompt Curriculum Learning for Efficient {LLM} Post-Training},
  author={Gao, Zhaolin and Kim, Joongwon and Sun, Wen and Joachims, Thorsten and Wang, Sid and Pang, Richard Yuanzhe and Tan, Liang},
  booktitle={International Conference on Learning Representations},
  year={2026}
}

@inproceedings{muennighoff2025s1,
  title = {s1: Simple test-time scaling},
  author = {Muennighoff, Niklas and Yang, Zitong and Shi, Weijia and Li, Xiang Lisa and Fei-Fei, Li and Hajishirzi, Hannaneh and Zettlemoyer, Luke and Liang, Percy and Cand{\`e}s, Emmanuel and Hashimoto, Tatsunori},
  booktitle = {Proceedings of the 2025 Conference on Empirical Methods in Natural Language Processing},
  pages = {20275--20321},
  year = {2025},
  address = {Suzhou, China},
  publisher = {Association for Computational Linguistics},
}

@article{qu2025adarft,
  title = {Efficient Reinforcement Finetuning via Adaptive Curriculum Learning},
  author = {Shi, Taiwei and Wu, Yiyang and Song, Linxin and Zhou, Tianyi and Zhao, Jieyu},
  journal = {Transactions on Machine Learning Research},
  year = {2026},
}

@inproceedings{qiu2025mopps,
  title = {Can Prompt Difficulty be Online Predicted for Accelerating {RL} Finetuning of Reasoning Models?},
  author = {Qu, Yun and Wang, Qi and Mao, Yixiu and Hu, Vincent Tao and Ommer, Bj{\"o}rn and Ji, Xiangyang},
  booktitle = {Proceedings of the 32nd ACM SIGKDD Conference on Knowledge Discovery and Data Mining},
  year = {2026}
}

@inproceedings{shen2025bots,
  title = {{BOTS}: A Unified Framework for {B}ayesian Online Task Selection in {LLM} Reinforcement Finetuning},
  author = {Shen, Qianli and Chen, Daoyuan and Huang, Yilun and Ling, Zhenqing and Li, Yaliang and Ding, Bolin and Zhou, Jingren},
  booktitle = {International Conference on Learning Representations},
  year = {2026},
}

@article{wu2026hive,
  title = {Train at Moving Edge: Online-Verified Prompt Selection for Efficient {RL} Training of Large Reasoning Model},
  author = {Wu, Jiahao and Lu, Ning and Liu, Shengcai and Wang, Kun and Yang, Yanting and Lin, Baijiong and Zhang, Chen Jason and {Li Qing} and Tang, Ke},
  journal = {arXiv preprint arXiv:2603.25184},
  year = {2026},
}

@inproceedings{ash2020badge,
  title={Deep Batch Active Learning by Diverse, Uncertain Gradient Lower Bounds},
  author={Ash, Jordan T. and Zhang, Chicheng and Krishnamurthy, Akshay and Langford, John and Agarwal, Alekh},
  booktitle={International Conference on Learning Representations},
  year={2020}
}

@inproceedings{kirsch2019batchbald,
  title={{BatchBALD}: Efficient and Diverse Batch Acquisition for Deep {B}ayesian Active Learning},
  author={Kirsch, Andreas and van Amersfoort, Joost and Gal, Yarin},
  booktitle={Advances in Neural Information Processing Systems},
  volume={32},
  year={2019}
}

@inproceedings{citovsky2021clustermargin,
  title={Batch Active Learning at Scale},
  author={Citovsky, Gui and DeSalvo, Giulia and Gentile, Claudio and Karydas, Lazaros and Rajagopalan, Anand and Rostamizadeh, Afshin and Kumar, Sanjiv},
  booktitle={Advances in Neural Information Processing Systems},
  volume={34},
  year={2021}
}

@article{wachi2026relativebudget,
  title={A Relative-Budget Theory for Reinforcement Learning with Verifiable Rewards in Large Language Model Reasoning},
  author={Wachi, Akifumi and Kinoshita, Hirota and Takakura, Shokichi and Higuchi, Rei and Suzuki, Taiji},
  journal={arXiv preprint arXiv:2602.01523},
  year={2026}
}

@inproceedings{sheng2025hybridflow,
  title={{HybridFlow}: A Flexible and Efficient {RLHF} Framework},
  author={Sheng, Guangming and Zhang, Chi and Ye, Zilingfeng and Wu, Xibin and Zhang, Wang and Zhang, Ru and Peng, Yanghua and Lin, Haibin and Wu, Chuan},
  booktitle={Proceedings of the Twentieth European Conference on Computer Systems},
  pages={1279--1297},
  year={2025}
}

@article{hu2025vade,
  title = {{VADE}: Variance-Aware Dynamic Sampling via Online Sample-Level Difficulty Estimation for Multimodal {RL}},
  author = {Hu, Zengjie and Qiu, Jiantao and Bai, Tianyi and Yang, Haojin and Yuan, Binhang and Jing, Qi and He, Conghui and Zhang, Wentao},
  journal = {arXiv preprint arXiv:2511.18902},
  year = {2025},
}

@article{leng2025mmr1,
  title = {{MMR1}: Enhancing Multimodal Reasoning with Variance-Aware Sampling and Open Resources},
  author = {Leng, Sicong and Wang, Jing and Li, Jiaxi and Zhang, Hao and Hu, Zhiqiang and Zhang, Boqiang and Jiang, Yuming and Zhang, Hang and Li, Xin and Bing, Lidong and Zhao, Deli and Lu, Wei and Rong, Yu and Sun, Aixin and Lu, Shijian},
  journal = {arXiv preprint arXiv:2509.21268},
  year = {2025},
}

@inproceedings{wang2025papo,
  title = {Perception-Aware Policy Optimization for Multimodal Reasoning},
  author = {Wang, Zhenhailong and Guo, Xuehang and Stoica, Sofia and Xu, Haiyang and Wang, Hongru and Ha, Hyeonjeong and Chen, Xiusi and Chen, Yangyi and Yan, Ming and Huang, Fei and Ji, Heng},
  booktitle = {International Conference on Learning Representations},
  year = {2026}
}

@inproceedings{liu2025mssr,
  title = {Stable and Efficient Single-Rollout {RL} for Multimodal Reasoning},
  author = {Liu, Rui and Yu, Dian and Ke, Lei and Liu, Haolin and Zhou, Yujun and Liang, Zhenwen and Mi, Haitao and Tokekar, Pratap and Yu, Dong},
  booktitle = {Proceedings of the IEEE/CVF Conference on Computer Vision and Pattern Recognition},
  year = {2026}
}

@inproceedings{huang2026calibrl,
  title = {Controllable Exploration in Hybrid-Policy {RLVR} for Multi-Modal Reasoning},
  author = {Huang, Zhuoxu and Jia, Mengxi and Sun, Hao and Li, Xuelong and Han, Jungong},
  booktitle = {International Conference on Learning Representations},
  year = {2026},
}

@inproceedings{liu2025noisyrollout,
  title = {{NoisyRollout}: Reinforcing Visual Reasoning with Data Augmentation},
  author = {Liu, Xiangyan and Ni, Jinjie and Wu, Zijian and Du, Chao and Dou, Longxu and Wang, Haonan and Pang, Tianyu and Shieh, Michael Qizhe},
  booktitle = {Advances in Neural Information Processing Systems},
  volume = {38},
  year = {2025}
}

\appendix

\section{Additional visual-dependence proxy evidence}
\label{app:visual-dependence-proxies}

\begin{figure}[!h]
\centering
\includegraphics[width=0.78\linewidth]{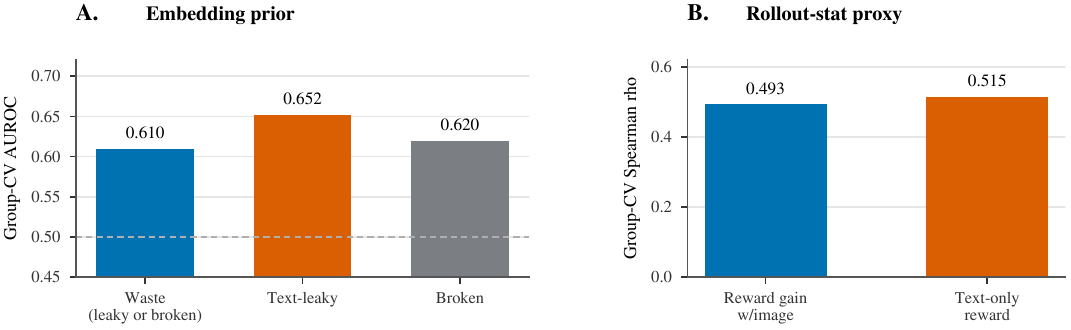}
\caption{Proxy evidence for visual dependence. (A) The embedding-prior
bars report group-CV AUROC for predicting \emph{No visual signal},
\emph{Text-only solved}, and \emph{Unsolved} labels from frozen prompt
embeddings, showing above-chance transfer to held-out clusters (AUROC
$0.610$--$0.652$). (B) The rollout-stat bars report group-CV Spearman
correlation for predicting image reward gain and text-only reward from
online rollout statistics, showing moderate transferable rank signal
($\rho{=}0.493$ and $0.515$).}
\label{fig:visual-dependence-proxies}
\end{figure}

Figure~\ref{fig:visual-dependence-proxies} checks whether the
audit-derived visual-dependence signal is recoverable from proxies that
are available during selection. Both panels use group cross-validation
by embedding cluster, so the reported transfer is measured on held-out
semantic neighborhoods rather than on prompts from the same cluster.
The \emph{No visual signal} label pools prompts that are solved without
the image or remain unsolved, since neither provides positive evidence
of image dependence.

\section{Full Selection Algorithm}
\label{app:algorithm}

\begin{algorithm}[t]
\caption{\method{}: budgeted online discovery for GRPO}
\label{alg:discover}
\small
\begin{algorithmic}[1]
\Require Prompt pool $\mathcal{D}$; verifier $r$; initial policy $\pi_0$; group size $G$;
budget $B$; clusters $K$; representatives $R$; schedule $\{(\kappa_p,\Delta_p)\}_{p=1}^{3}$;
KNN size $k$; decay $\lambda$.
\Ensure Trained policy $\pi_T$; verified set $\mathcal{V}_{\le T}$; replay pool $\mathcal{P}$;
rollout buffer $\mathcal{H}_T$.

\Statex \textbf{Offline organization}
\State Embed all prompts with a frozen multimodal encoder and normalize to $z_i$.
\State Cluster $\{z_i\}$ into $C_1,\ldots,C_K$ with spherical $k$-means.
\State Form $\mathcal{M}$ by greedily selecting up to $R$ prompts per cluster using
$g_i=\tilde d_i s_i$ and the $0.85$ similarity cutoff, with the small-cluster and fallback
rules described below.

\Statex \textbf{Initialization}
\State $\mathcal{H}_0\gets\emptyset$, $\mathcal{V}_{\le 0}\gets\mathcal{M}$,
$\mathcal{R}_0\gets\mathcal{M}$, $\mathcal{T}_{\le 0}\gets\emptyset$, $\phi\gets 0$, $\pi\gets\pi_0$.
\ForAll{$m\in\mathcal{M}$}
    \State Generate $G$ rollouts with $\pi_0$, verify correctness, and append
    $(m,0,\{\mathrm{acc}_{m,g}\}_{g=1}^{G})$ to $\mathcal{H}_0$.
\EndFor

\Statex \textbf{Online discovery and frozen reweighting}
\For{$t=0,\ldots,T-1$}
    \State If $t>0$, carry forward $\mathcal{V}$, $\mathcal{T}$, $\mathcal{H}$, and $q$ from step $t-1$.
    \State Set $\mathcal{V}_{<t}\gets\mathcal{V}_{\le t}$ and the phase $p$ from
    $f_t=|\mathcal{V}_{<t}|/B$ ($p=3$ once frozen).
    \If{$t \bmod \Delta_p = 0$} \Comment{includes $t=0$}
        \State Compute $(\bar\mu_j(t),\bar v_j(t))$ from $\mathcal{H}_t$ with weights
        $e^{-\lambda(t-\tau)}$; let $\mathcal{R}_t$ be the verified prompts with valid statistics.
        \State Transfer $(\hat\mu_i(t),\hat v_i(t))$ to all $i\in\mathcal{D}$ by cosine-KNN over
        $\mathcal{R}_t$, and compute $u_i(t)=\hat v_i(t)\,a(\hat\mu_i(t))\,m_{\mathrm{dz}}(\hat\mu_i(t))$.
        \If{$\phi=0$}
            \State $\mathcal{U}_t\gets\mathcal{D}\setminus\mathcal{V}_{<t}$;\;
            $\bar b_t\gets\min(\lfloor\kappa_p N_{\mathrm{tr}}\rfloor,\,B-|\mathcal{V}_{<t}|,\,|\mathcal{U}_t|)$.
            \State Set $E_c(t)\gets C_c\cap\mathcal{U}_t$ and $b_c(t)\gets 0$ for every cluster;
            for nonempty $E_c(t)$, compute $A_c(t)=|E_c(t)|^{-1}\sum_{i\in E_c(t)}u_i(t)$ and
            retain clusters with $A_c(t)>0$.
            \If{at least one cluster is retained}
                \State $b_c(t)\gets\mathrm{round}(\bar b_t\rho_c(t))$ for retained clusters, with
                $\rho_c(t)\propto\exp(A_c(t)/T_c)$; repair within capacities using the rules below.
            \EndIf
            \State $\mathcal{A}_t\gets\bigcup_c\operatorname{Top}_{b_c(t)}(E_c(t);u(t))$; add the
            highest-utility prompts in $\mathcal{U}_t\setminus\mathcal{A}_t$ until $|\mathcal{A}_t|=\bar b_t$.
            \State Verify $\mathcal{A}_t$ and add it to $\mathcal{V}_{\le t}$ and $\mathcal{T}_{\le t}$
            (Training-first).
            \State Admit verified prompts in $\mathcal{V}_{\le t}\setminus\mathcal{T}_{\le t}$
            (e.g., REPR) to $\mathcal{T}_{\le t}$ by utility.
            \If{$|\mathcal{V}_{\le t}|=B$}
                \State $\phi\gets 1$ and freeze $\mathcal{P}\gets\mathcal{T}_{\le t}$.
            \EndIf
        \EndIf
        \If{$\phi=1$}
            \State Update $q_t$ over $\mathcal{P}$ from $u_i(t)$ with the starvation floor
            (Eq.~\ref{eq:reweight}).
        \EndIf
    \EndIf
    \State Sample a GRPO batch uniformly from $\mathcal{T}_{\le t}$ if $\phi=0$, otherwise from
    $\mathcal{P}$ using $q_t$.
    \State Generate $G$ rollouts per prompt; update $\pi$ with
    $0.9\times$correctness$\,+\,0.1\times$format; append correctness to $\mathcal{H}_t$.
\EndFor
\State \Return $\pi_T,\mathcal{V}_{\le T},\mathcal{P},\mathcal{H}_T$.
\end{algorithmic}
\end{algorithm}

\begin{table}[h]
\centering
\caption{Prompt-set notation used in Section~3 and Algorithm~\ref{alg:discover}.}
\label{tab:symbols}
\small
\begin{tabular}{ll}
\toprule
\textbf{Symbol} & \textbf{Definition} \\
\midrule
$\mathcal{D}$ & Full candidate prompt pool \\
$\mathcal{M}$ & Initial REPR prompts \\
$\mathcal{V}_{\le t}$ & Unique prompts verified up to step $t$ (bounded by budget $B$) \\
$\mathcal{V}_{<t}$ & Verified set before discovery at step $t$, including initial REPR queries \\
$\mathcal{U}_t$ & Available unverified prompts ($\mathcal{D}\setminus\mathcal{V}_{<t}$) \\
$\mathcal{R}_t$ & Verified prompts used for KNN transfer ($\mathcal{R}_t\subseteq\mathcal{V}_{\le t}$) \\
$\mathcal{T}_{\le t}$ & Verified prompts admitted to the GRPO pool by step $t$ ($\mathcal{T}_{\le t}\subseteq\mathcal{V}_{\le t}$) \\
$\mathcal{P}$ & Frozen replay pool used after budget exhaustion \\
\bottomrule
\end{tabular}
\end{table}

Algorithm~\ref{alg:discover} maintains $\mathcal{V}_{\le t}$, $\mathcal{R}_t$,
and $\mathcal{T}_{\le t}$ separately (Table~\ref{tab:symbols}). Initial REPR
queries enter $\mathcal{V}_{\le 0}$ and initialize $\mathcal{R}_0$, while
$\mathcal{T}_{\le 0}$ is empty. Their later admission to $\mathcal{T}_{\le t}$
follows the utility rule without increasing the verified-prompt count, because
the budget counts unique prompts. At the end of discovery,
$\mathcal{T}_{\le T}\subseteq\mathcal{V}_{\le T}$ and
$\mathcal{P}=\mathcal{T}_{\le T}$; after budget exhaustion, replay draws from
$\mathcal{P}$. The final Probe-only set is
$\mathcal{V}_{\le T}\setminus\mathcal{T}_{\le T}$.

The ablations in Table~\ref{tab:components} correspond to simple switches in
this loop. Cluster-only removes the KNN transfer step and allocates by cluster
coverage alone. Immediate freeze performs one discovery pass and then sets $\phi=1$. Random-score replaces $u_i(t)$ with random scores under the same
budget, with either the default or the slow schedule. MLP and Ridge replace the nonparametric KNN transfer rule
with learned regressors while keeping the remaining discovery loop fixed.

\paragraph{REPR selection.}
Before training, we select up to $R$ representative prompts (REPR)
from each cluster to obtain the selector's initial reward
observations. This procedure is performed once, with clusters
containing at most $R$ prompts contributing all their prompts
without scoring.

For larger clusters, candidates are scored by local density and
proximity to the cluster centroid. Since embeddings are
$L_2$-normalized, $z_i^\top z_j$ is cosine similarity. Let
$\mathrm{NN}_k(i)$ contain the $k=\min(10,|C_c|-1)$ nearest
in-cluster neighbours of candidate $i$, excluding $i$ itself.
We compute local density as
\[
d_i=\left(
\frac{1}{k}\sum_{j\in\mathrm{NN}_k(i)}
(1-z_i^\top z_j)+\epsilon
\right)^{-1},
\qquad \epsilon=10^{-8},
\]
and normalize it within the cluster to obtain
$\tilde d_i\in[0,1]$. We combine this normalized density with
centroid proximity
$s_i=\operatorname{clip}_{[0,1]}(z_i^\top\hat\mu_c)$
to obtain the fixed candidate score $g_i=\tilde d_i s_i$,
where $\hat\mu_c$ is the normalized cluster centroid.

We greedily select up to $R$ prompts per cluster, repeatedly
taking the highest-scoring remaining candidate $i^\star$ and
excluding candidates $j$ with $z_j^\top z_{i^\star}>0.85$
to suppress near-duplicates. If all remaining candidates have
$g_i\le0$, we rank them by raw density instead. Ties are broken
by fixed candidate index, and initialization selects at most
$KR$ prompts across all clusters.

\paragraph{Cluster allocation.}
The rounded allocations $\mathrm{round}(\bar b_t\rho_c(t))$
are capped at each cluster's available capacity $|E_c(t)|$.
When the positive-utility clusters have sufficient capacity,
we repair rounding to satisfy $\sum_c b_c(t)=\bar b_t$.
Surplus is removed from the lowest-scoring clusters, while
deficits are assigned to the highest-scoring clusters with
remaining capacity, with ties broken by cluster index.
If no cluster has positive utility, or the positive-utility
clusters cannot fill the quota, remaining slots are filled
with the globally highest-utility unverified prompts not
already selected.

\paragraph{Cadence.}
Discovery is triggered at absolute optimizer steps, so a phase
change can make the gap between consecutive rounds shorter
than the newly active interval. After freezing, replay weights
continue to update at the final phase's interval, using the
same absolute-step trigger.
On ViRL39K ($N_{\rm tr}=34{,}983$), the headline phases request
$349$, $104$, and $52$ prompts per round before clipping, and the
initial REPR queries use about $11.6\%$ of the $10\%$ budget.

\subsection{Selector Configuration}
\label{app:selector-config}

\method{} uses frozen Qwen3-VL-Embedding-2B~\citep{li2026qwen3vlembedding} embeddings with $K{=}150,R{=}3$ on ViRL39K and $K{=}200,R{=}2$ on VLAA-Thinking, selected through inertia-elbow and silhouette sweeps (Appendix~\ref{app:embedding-clusters}).

We denote schedules as $(100\kappa_p,\Delta_p)$ for $p=1,2,3$. On ViRL39K, the headline schedule uses $(1.0,4){\to}(0.30,10){\to}(0.15,16)$, and the slow control in Table~\ref{tab:components} uses $(0.70,5){\to}(0.25,15){\to}(0.10,25)$. The headline configuration combines semantic coverage, KNN reward-statistic transfer, online refresh, and asymmetric mean-aware utility. In Table~\ref{tab:components}, one-shot denotes a single initial discovery pass followed by a frozen training pool.

\section{Embedding, clustering, and utility-transfer analyses}
\label{app:embedding-clusters}

\begin{figure}[t]
\centering
\includegraphics[width=0.88\linewidth]{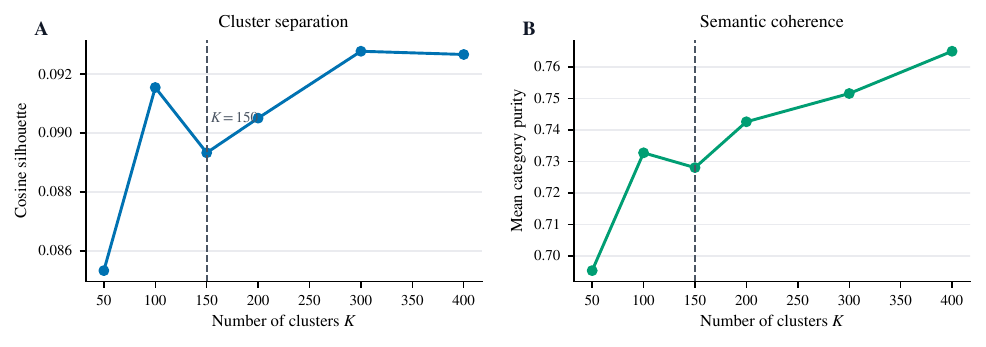}
\caption{ViRL39K cluster-count sweep on frozen Qwen3-VL embeddings. The dashed
line marks the selected operating point $K=150$. Cosine silhouette and
dominant-category purity both show that the embedding geometry groups prompts
into coherent neighborhoods, while $K=150$ keeps representative-probe overhead
small.}
\label{fig:cluster-sweep}
\end{figure}

We build the selector geometry from the ViRL39K train split using
$34{,}983$ Qwen3-VL prompt embeddings of dimension $2048$. The cached
embeddings are already $L^2$ normalized (mean norm $1.000$, standard deviation
$1.7{\times}10^{-4}$). Figure~\ref{fig:cluster-sweep} reports the cluster-count
sweep used to choose the operating point. The sweep identifies $K=150$ as the
elbow and recommended value: it keeps the representative-probe set small
($450$ prompts, $1.16\%$ of the full $38{,}870$-prompt source pool) while
preserving semantic coherence. At $K=150$, the cosine silhouette is $0.089$ and
the mean dominant-category purity is $0.728$; larger $K$ values slightly
increase purity but also increase the number of representative probes.

\begin{figure}[t]
\centering
\includegraphics[width=0.88\linewidth]{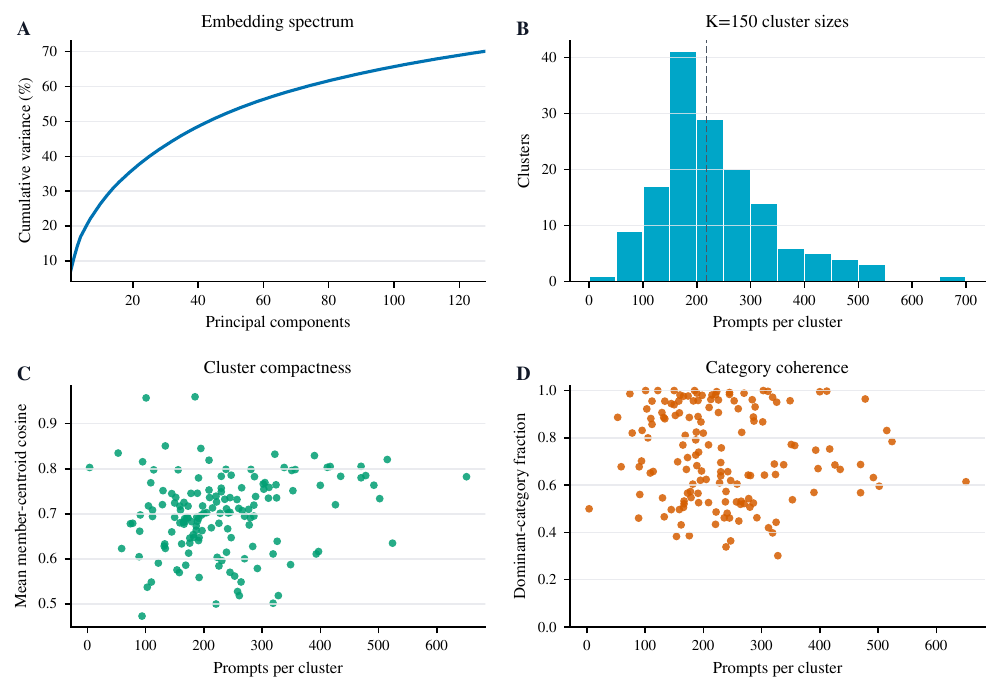}
\caption{Embedding and cluster statistics for the selected ViRL39K geometry.
(A) The frozen embedding spectrum is broad rather than dominated by one
direction. (B) The $K=150$ partition gives moderately balanced nonempty
clusters. (C) Member-centroid cosine values show compact semantic
neighborhoods. (D) Dominant-category fractions show that many clusters align
with dataset-level semantic categories.}
\label{fig:embedding-cluster-analysis}
\end{figure}

Figure~\ref{fig:embedding-cluster-analysis} summarizes the selected
$K=150$ geometry. All clusters are nonempty. The mean cluster size is $233.2$
prompts, with median $218$, 10th--90th percentile range $111.8$--$390.3$, and
Gini coefficient $0.248$. The embedding spectrum is moderately high-dimensional:
the first principal component explains $7.20\%$ of variance, the first $64$
components explain $57.5\%$, the first $128$ explain $70.1\%$, and the
participation ratio is $80.8$. This supports using a neighborhood-based
transfer rule rather than a one-dimensional score or a small number of coarse
dataset partitions.

\paragraph{Spherical versus standard $k$-means.}
On a $6{,}000$-prompt ViRL39K subsample with $K=150$, standard
$k$-means centroids have a mean norm of $0.715$, whereas spherical
$k$-means centroids have unit norm. The two algorithms
assign only $0.93\%$ of prompts to different clusters.

\paragraph{Clusters and difficulty.}
We estimate prompt difficulty from eight base-model rollouts on a random
sample of ViRL39K prompts. Relative to 500 shuffled assignments that preserve
cluster sizes, clustering reduces within-cluster difficulty variation by
$7.8\%$ ($p<0.002$). The reduction ranges from $5.8\%$ to $10.7\%$ for
$K\in\{50,100,150,200,300\}$. Among clusters with at least eight audited
prompts, $76.8\%$ contain both fully solved and fully failed prompts.

\paragraph{Held-out utility transfer.}
The held-out set is excluded from GRPO training and from the selector's reward
observations in the reported run. For each prompt, we predict utility from its
$k{=}64$ nearest cosine neighbors and compare the prediction with utility computed
from the prompt's own rollouts. Table~\ref{tab:heldout-utility-transfer}
reports the Spearman correlation and the top-$10\%$ lift, defined as the
percentage increase in mean realized utility among the highest-ranked $10\%$
relative to the held-out-set mean. At the \method{} checkpoint, $73.0\%$
of the top-ranked $10\%$ have non-uniform correctness across rollouts, compared
with $48.8\%$ of all held-out prompts. From the base model to \method{}, the
fraction of prompts with uniform correctness rises from $45.0\%$ to $51.2\%$,
and neighbor-based prediction of uniform outcomes improves from AUROC $0.568$
to $0.700$.

\paragraph{Alternative representations.}
At the \method{} checkpoint, we repeat the analysis with policy-backbone
features of the pre-RL model, random Gaussian vectors, PCA projections, and
embeddings with additive Gaussian noise. Utility transfer holds across
representations: policy-backbone features also predict realized utility
($\rho{=}0.233$), whereas random vectors and random neighbors give correlations
near zero. The signal is robust to compression and perturbation, with $\rho$
between $0.29$ and $0.36$ from 128 down to 16 PCA dimensions and $\rho{=}0.34$
under additive Gaussian noise. No GRPO run used an alternative representation.

\begin{table}[t]
\centering
\caption{Held-out utility transfer on ViRL39K. $\rho$ is the Spearman
correlation between predicted and realized utility; lift is the percentage
increase in mean realized utility among the highest-ranked $10\%$ relative to
the held-out-set mean. Brackets are $95\%$ bootstrap intervals over held-out
prompts; random-vector and random-neighbor entries are mean $\pm$ SD over seeds
or draws, with 2.5--97.5 percentile ranges. Policy-backbone features are
evaluated on prompts for which they are available. Noise is additive Gaussian
at the per-coordinate scale.}
\label{tab:heldout-utility-transfer}
\small
\begin{adjustbox}{max width=\textwidth}
\begin{tabular}{llcc}
\toprule
Checkpoint & Representation / neighbor rule & $\rho$ & Top-$10\%$ lift \\
\midrule
Base & Frozen encoder & 0.163 [0.068, 0.259] & $+38\%$ [$5\%$, $61\%$] \\
\method{}-$10\%$ & Frozen encoder & 0.364 [0.273, 0.449] & $+48\%$ [$17\%$, $80\%$] \\
\method{}-$10\%$ & Frozen encoder, $k=16$ & 0.362 [0.268, 0.448] & $+49\%$ [$17\%$, $86\%$] \\
\method{}-$10\%$ & Policy-backbone features (pre-RL) & 0.233 [0.127, 0.331] & $+36\%$ [$0\%$, $69\%$] \\
\method{}-$10\%$ & PCA, 128 dims ($70\%$ var.) & 0.358 [0.266, 0.445] & $+39\%$ [$12\%$, $74\%$] \\
\method{}-$10\%$ & PCA, 32 dims ($44\%$ var.) & 0.310 [0.212, 0.401] & $+35\%$ [$8\%$, $72\%$] \\
\method{}-$10\%$ & PCA, 16 dims ($33\%$ var.) & 0.288 [0.189, 0.381] & $+53\%$ [$19\%$, $81\%$] \\
\method{}-$10\%$ & Additive Gaussian noise & 0.340 [0.247, 0.426] & $+57\%$ [$28\%$, $88\%$] \\
\midrule
\method{}-$10\%$ & Random Gaussian vectors & $-0.020\pm0.076$ [$-0.120$, $0.122$] & $-1\%\pm23\%$ [$-42\%$, $36\%$] \\
\method{}-$10\%$ & Random neighbors & $-0.018\pm0.050$ [$-0.117$, $0.075$] & $-5\%\pm17\%$ [$-31\%$, $30\%$] \\
\bottomrule
\end{tabular}
\end{adjustbox}
\end{table}

\begin{table}[!h]
\centering
\caption{In-buffer transfer diagnostics on ViRL39K. Leave-one-out
Spearman within the selector buffer, using prior observations and excluding
the target prompt; mean $\pm$ SD over selection rounds.}
\label{tab:selector-transfer}
\small
\begin{adjustbox}{max width=\textwidth}
\begin{tabular}{llc}
\toprule
\textbf{Variant} & \textbf{Selector rule} & \textbf{LOO Spearman} \\
\midrule
KNN transfer & KNN & $0.264 \pm 0.063$ \\
MLP transfer & MLP & $0.182 \pm 0.054$ \\
Ridge transfer & Ridge & $0.142 \pm 0.051$ \\
Ridge $+$ active probes & Ridge+AP &  $0.163 \pm 0.033$ \\
Random-score slow & random & $0.103 \pm 0.015$ \\
\bottomrule
\end{tabular}
\end{adjustbox}
\end{table}

\section{Bootstrap uncertainty over benchmark items}
\label{app:benchmark-bootstrap}

We estimate evaluation uncertainty by nonparametric bootstrap over benchmark
items. For each evaluated model, we resample items within each benchmark,
compute that benchmark's accuracy, and average the four benchmark accuracies;
the reported interval is the 2.5--97.5 percentile range over 5{,}000 bootstrap
replicates. These intervals quantify evaluation-set uncertainty for the
evaluated models. Because the models are evaluated on the same benchmark items,
we also report paired-bootstrap intervals for \method{}'s gain over each
evaluated baseline. Intervals are benchmark-stratified over 10{,}994 matched evaluation items.

\begin{table}[t]
\centering
\caption{Bootstrap uncertainty over external benchmark items for evaluated
ViRL39K models. Intervals are 95\% nonparametric CIs for the four-benchmark
average.}
\label{tab:benchmark-bootstrap-ci}
\scriptsize
\setlength{\tabcolsep}{5pt}
\begin{adjustbox}{max width=0.68\textwidth}
\begin{tabular}{lcc}
\toprule
\textbf{Method} & \textbf{Avg. $\uparrow$} & \textbf{95\% CI} \\
\midrule
\method{}-$10\%$ & \textbf{13.12} & [12.21, 14.07] \\
MLP transfer & 11.32 & [10.46, 12.27] \\
Random score (slow) & 10.76 & [9.96, 11.62] \\
One-shot KNN & 10.03 & [9.20, 10.93] \\
Slow cadence & 9.51 & [8.73, 10.32] \\
DOTS-MM-$10\%$ & 9.12 & [8.32, 10.00] \\
VADE original & 8.32 & [7.58, 9.13] \\
Coverage only & 8.23 & [7.46, 9.06] \\
\bottomrule
\end{tabular}
\end{adjustbox}
\end{table}

\begin{table}[t]
\centering
\caption{Paired bootstrap gains over evaluated ViRL39K baselines. Each
interval resamples the same benchmark items for \method{} and the baseline, then
recomputes the four-benchmark average difference in percentage points.}
\label{tab:benchmark-paired-bootstrap-ci}
\scriptsize
\setlength{\tabcolsep}{5pt}
\begin{adjustbox}{max width=0.82\textwidth}
\begin{tabular}{lcc}
\toprule
\textbf{Comparison} & \textbf{$\Delta$ Avg. (pp) $\uparrow$} & \textbf{95\% CI} \\
\midrule
\method{}-$10\%$ vs. Random-$10\%$ & 2.74 & [1.95, 3.53] \\
\method{}-$10\%$ vs. MaxVar-$10\%$ & 2.08 & [1.01, 3.12] \\
\method{}-$10\%$ vs. Hard-$10\%$ & 4.38 & [3.50, 5.24] \\
\method{}-$10\%$ vs. Full train split & 3.41 & [2.52, 4.30] \\
\midrule
\method{}-$10\%$ vs. MLP transfer & 1.80 & [0.85, 2.73] \\
\method{}-$10\%$ vs. Random score (slow) & 2.35 & [1.41, 3.33] \\
\method{}-$10\%$ vs. One-shot KNN & 3.08 & [2.11, 4.06] \\
\method{}-$10\%$ vs. Slow cadence & 3.61 & [2.81, 4.42] \\
\method{}-$10\%$ vs. DOTS-MM-$10\%$ & 3.99 & [3.14, 4.83] \\
\method{}-$10\%$ vs. VADE original & 4.79 & [3.96, 5.66] \\
\method{}-$10\%$ vs. Coverage only & 4.88 & [4.07, 5.75] \\
\bottomrule
\end{tabular}
\end{adjustbox}
\end{table}

\section{Evaluation with a larger generation cap}
\label{app:eval-2048}

Tables~\ref{tab:main-bench-2048} and~\ref{tab:components-2048} repeat the
comparisons in Tables~\ref{tab:main-bench} and~\ref{tab:components} with a
2{,}048-token generation cap. All other evaluation settings are unchanged. All
ablations remain below the full method; the gaps for MLP transfer and slow
random scoring are smaller than with the 512-token cap.

\begin{table}[t]
\centering
\caption{External benchmark accuracy (\%) with a 2{,}048-token generation cap.
Exposure and selector-compute columns are as in Table~\ref{tab:main-bench}.
Bold marks the best result per dataset.}
\label{tab:main-bench-2048}
\scriptsize
\setlength{\tabcolsep}{1.6pt}
\renewcommand{\arraystretch}{1.12}
\begin{adjustbox}{max width=\textwidth}
\begin{tabular}{>{\centering\arraybackslash}m{0.72cm} l!{\vrule width 0.25pt} c c c c c !{\vrule width 0.25pt} c !{\vrule width 0.25pt} c c c c c}
\toprule
\textbf{Dataset} & \textbf{Method}
& \multicolumn{5}{!{\vrule width 0.25pt}c!{\vrule width 0.25pt}}{\textbf{Verified-prompt exposure (\%) $\downarrow$}}
& \multicolumn{1}{c!{\vrule width 0.25pt}}{\textbf{Selector compute $\downarrow$}}
& \multicolumn{5}{c}{\textbf{External benchmark accuracy (\%) $\uparrow$}} \\
\cmidrule(lr){3-7}\cmidrule(lr){8-8}\cmidrule(lr){9-13}
& & \multicolumn{2}{c}{\textbf{By acquisition}} & \multicolumn{2}{c}{\textbf{By final use}} & & & & & & & \\
\cmidrule(lr){3-4}\cmidrule(lr){5-6}
& & \textbf{Selector-} & \textbf{Training-} & \textbf{Train} & \textbf{Probe-} & \textbf{Total}
& \textbf{Probe Gen.}
& \textbf{WeMath} & \textbf{MathVerse} & \textbf{MATH-Vision} & \textbf{DynaMath} & \textbf{Avg.} \\
& & \textbf{first} & \textbf{first} & & \textbf{only} &
& \textbf{($\times N$)}
& & & & & \\
\midrule
Base & Qwen3-VL-2B
& -- & -- & -- & -- & -- & --
& 3.05 & 9.59 & 7.89 & 19.46 & 10.00 \\
\midrule
\multirow{10}{*}{\rotatebox[origin=c]{90}{ViRL39K \citep{wang2025virl}}}
& Random-$5\%$
& 0.00 & 5.00 & 5.00 & 0.00 & 5.00 & 0
& 16.26 & 11.37 & 23.03 & 23.89 & 18.64 \\
& Random-$10\%$
& 0.00 & 10.00 & 10.00 & 0.00 & 10.00 & 0
& 19.25 & 11.55 & 20.07 & 22.87 & 18.44 \\
& Hard-$10\%$
& 90.00 & 0.00 & 10.00 & 80.00 & 90.00 & 0.90
& 6.15 & 12.51 & 18.75 & 22.16 & 14.89 \\
& MaxVar-$10\%$
& 90.00 & 0.00 & 10.00 & 80.00 & 90.00 & 7.20
& 13.10 & 11.62 & \textbf{25.00} & 21.78 & 17.88 \\
& DOTS-MM-$10\%$ \citep{sun2025dots}
& 3.36 & 6.64 & \rng{6.64,10.00}{9.68} & \rng{0.00,3.36}{0.32} & 10.00 & 0.27
& 7.82 & 10.10 & 22.70 & 22.38 & 15.75 \\
& VADE original \citep{hu2025vade}
& 90.00 & 0.00 & 47.77 & 42.23 & 90.00 & 7.20
& 6.72 & 11.50 & 18.09 & 22.22 & 14.63 \\
& VADE-$10\%$ \citep{hu2025vade}
& 90.00 & 0.00 & 10.00 & 80.00 & 90.00 & 7.20
& 14.31 & 11.55 & 18.75 & 23.03 & 16.91 \\
& Full train split
& 0.00 & 90.00 & 90.00 & 0.00 & 90.00 & 0
& 15.34 & 13.07 & 16.12 & 23.33 & 16.97 \\
\cmidrule(lr){2-13}
& \method{}-$5\%$
& 1.16 & 3.84 & \rng{3.84,5.00}{4.49} & \rng{0.00,1.16}{0.51} & 5.00 & 0.09
& 20.93 & 13.28 & 18.21 & 23.94 & 19.09 \\
& \textbf{\method{}-$10\%$}
& 1.16 & 8.84 & \rng{8.84,10.00}{9.57} & \rng{0.00,1.16}{0.43} & 10.00 & 0.09
& \textbf{26.21} & \textbf{13.40} & 18.42 & \textbf{27.31} & \textbf{21.33} \\
\midrule
\multirow{10}{*}{\rotatebox[origin=c]{90}{VLAA-Thinking \citep{chen2504vlaa}}}
& Random-$5\%$
& 0.00 & 5.00 & 5.00 & 0.00 & 5.00 & 0
& 22.76 & 17.28 & 10.86 & 27.47 & 19.59 \\
& Random-$10\%$
& 0.00 & 10.00 & 10.00 & 0.00 & 10.00 & 0
& 16.21 & 16.88 & 14.14 & 25.71 & 18.23 \\
& Hard-$10\%$
& 90.00 & 0.00 & 10.00 & 80.00 & 90.00 & 0.90
& 14.76 & 16.97 & 13.72 & 26.23 & 17.92 \\
& MaxVar-$10\%$
& 90.00 & 0.00 & 10.00 & 80.00 & 90.00 & 7.20
& 18.71 & 17.43 & 14.06 & 28.25 & 19.61 \\
& DOTS-MM-$10\%$
& 4.60 & 5.40 & \rng{5.40,10.00}{9.56} & \rng{0.00,4.60}{0.44} & 10.00 & 0.37
& 18.16 & 16.98 & 12.17 & 27.41 & 18.68 \\
& VADE original
& 90.00 & 0.00 & 47.25 & 42.75 & 90.00 & 7.20
& 11.14 & 16.63 & 13.08 & 27.48 & 17.08 \\
& VADE-$10\%$
& 90.00 & 0.00 & 10.00 & 80.00 & 90.00 & 7.20
& 17.16 & 16.27 & 13.07 & 26.25 & 18.19 \\
& Full train split
& 0.00 & 90.00 & 90.00 & 0.00 & 90.00 & 0
& 19.77 & 16.90 & 15.79 & 27.84 & 20.08 \\
\cmidrule(lr){2-13}
& \method{}-$5\%$
& 1.59 & 3.41 & \rng{3.41,5.00}{4.74} & \rng{0.00,1.59}{0.26} & 5.00 & 0.13
& 21.23 & 18.16 & 13.98 & 28.19 & 20.39 \\
& \textbf{\method{}-$10\%$}
& 1.59 & 8.41 & \rng{8.41,10.00}{9.80} & \rng{0.00,1.59}{0.20} & 10.00 & 0.13
& \textbf{28.74} & \textbf{20.00} & \textbf{16.12} & \textbf{29.00} & \textbf{23.46} \\
\bottomrule
\end{tabular}
\end{adjustbox}
\end{table}

\begin{table}[t]
\centering
\caption{Component ablations on ViRL39K with a 2{,}048-token generation cap.
Columns are as in Table~\ref{tab:components}. Each row reports one training run.}
\label{tab:components-2048}
\scriptsize
\setlength{\tabcolsep}{2.8pt}
\renewcommand{\arraystretch}{1.13}
\begin{adjustbox}{max width=\textwidth}
\begin{tabular}{c l l cccccc}
\toprule
\textbf{Group} & \textbf{Intervention} & \textbf{Tests} & \textbf{Transfer} & \textbf{Utility} & \textbf{Refresh} & \textbf{Cadence} & \textbf{Avg. $\uparrow$} & \textbf{Ent.} \\
\midrule
\multirow{1}{*}{Reference}
& Full \method{}
& Reference setting
& KNN & asym & \checkmark & default
& $\mathbf{21.33}$ & $0.20$ \\
\midrule
\multirow{7}{*}{Predictor}
& Coverage only
& Coverage without utility
& \xmark & \xmark & \xmark & one-shot
& $13.80$ & $0.40$ \\
& One-shot KNN
& One-shot vs. online KNN
& KNN & asym & \xmark & one-shot
& $16.47$ & $0.30$ \\
& MLP transfer
& Learned vs. local transfer
& MLP & asym & \checkmark & default
& $20.14$ & $0.23$ \\
& Ridge transfer
& Linear vs. local transfer
& Ridge & asym & \checkmark & default
& $14.94$ & $0.28$ \\
& Ridge active probes
& More probes for Ridge
& Ridge+AP & asym & \checkmark & default
& $18.08$ & $0.09$ \\
& Random score
& Random predictor
& random & asym & \checkmark & default
& $17.16$ & $0.26$ \\
& Random score (slow)
& Random predictor
& random & asym & \checkmark & slow
& $20.12$ & $0.36$ \\
\midrule
\multirow{2}{*}{Utility}
& No hard-side bias
& Asymmetric utility bias
& KNN & symmetric & \checkmark & default
& $17.89$ & $0.38$ \\
& Dead-zone $h{=}0.8$
& Easy-filter aggressiveness
& KNN & $h{=}0.8$ & \checkmark & default
& $16.15$ & $0.12$ \\
\midrule
\multirow{1}{*}{Cadence}
& Slow cadence
& Delayed-freeze control
& KNN & asym & \checkmark & slow
& $16.23$ & $0.28$ \\
\bottomrule
\end{tabular}
\end{adjustbox}
\end{table}

\section{Training-seed variation}
\label{app:seeds}

To measure run-to-run variation, we ran two additional subset seeds for \method{} and Random at the
5\% and 10\% budgets on ViRL39K. Seeds were fixed in advance, all launched runs are reported, and
evaluation is identical to Table~\ref{tab:main-bench}. Seed I is the run reported in
Table~\ref{tab:main-bench}. Table~\ref{tab:seeds} reports the four-benchmark average of each run.

At 10\%, \method{} outperforms Random in all three runs, with a mean paired margin of
$+3.25\pm0.45$ points and a 95\% paired-$t$ confidence interval of $[2.12, 4.37]$ points for the
mean run-level difference. At 5\%, the means are $12.93\pm2.29$ for \method{} and $9.96\pm1.35$ for
Random. The higher variability places the single-run 5\% reversal in Table~\ref{tab:main-bench}
within subset-seed noise. Results at the 5\% budget have higher variance across seeds than at 10\%. These intervals
quantify training-run variation and are reported separately from the evaluation-item bootstrap in
Appendix~\ref{app:benchmark-bootstrap}.

\begin{table}[h]
\centering
\caption{Training-seed variation on ViRL39K. Four-benchmark average accuracy (\%) over three subset
seeds. Seed I is the run in Table~\ref{tab:main-bench}. The last column reports mean $\pm$ standard
deviation over seeds.}
\label{tab:seeds}
\small
\begin{tabular}{lcccc}
\toprule
\textbf{Method} & \textbf{Seed I} & \textbf{Seed II} & \textbf{Seed III} & \textbf{Mean $\pm$ SD} \\
\midrule
\method{}-$10\%$ & 13.12 & 13.86 & 12.78 & $\mathbf{13.25\pm0.55}$ \\
Random-$10\%$ & 10.38 & 10.25 & 9.39 & $10.01\pm0.54$ \\
\midrule
\method{}-$5\%$ & 10.53 & 13.18 & 15.09 & $12.93\pm2.29$ \\
Random-$5\%$ & 10.94 & 8.42 & 10.52 & $9.96\pm1.35$ \\
\bottomrule
\end{tabular}
\end{table}

\section{Prompt-exposure and selector-cost accounting}
\label{app:prompt-exposure}

We report two cost axes: verified-prompt exposure and selector-generation cost. Verified-prompt
exposure counts unique prompts that receive verifier feedback. We distinguish how a prompt is first
verified from how it is ultimately used, which partitions the same verified set in two ways
(notation in Table~\ref{tab:symbols}):
\begin{itemize}
    \item \textbf{By acquisition.} \textbf{Selector-first} prompts are queried to initialize or
    score the selector (e.g., REPR); \textbf{Training-first} prompts are queried upon admission
    during online discovery.
    \item \textbf{By final use.} \textbf{Train} prompts are used for GRPO sampling/replay,
    regardless of acquisition route; \textbf{Probe-only} prompts are verified and scored but never
    admitted for training.
\end{itemize}
Selector-first$+$Training-first $=$ Train$+$Probe-only $=$ Total verified. All percentages are
relative to the full source pool ($N$ prompts). Probe Gen. reports selector-only generations used
for initialization or prompt scoring, rather than unique prompts, and therefore does not affect the
verified-prompt percentage; it is reported as a multiple of $N$.
Scanning the $90\%$ training split with one rollout per prompt costs $0.9N$ generations, and $G{=}8$ rollouts per prompt cost $7.2N$. On ViRL39K ($N{=}38{,}870$), these scans require $34{,}983$ generations for Hard-$10\%$ and $279{,}864$ for MaxVar-$10\%$ and both VADE variants; \method{} uses $450G{=}3{,}600$ initialization generations at either budget. VADE-original's Train and Probe-only percentages are converted from the training-split denominator to the full-source denominator by multiplying by $0.9$. Full-data GRPO runs for the same number of optimizer steps, so it visits each training prompt about once, whereas $10\%$-budget methods revisit each selected prompt about ten times. All GRPO-trained methods share the optimizer-step budget, rollout group size, and implementation, so RL compute is matched up to selector overhead.

Because 10\% of each source pool is held out for validation, dense-scored baselines scan the eligible
90\% training split before choosing a 10\% training subset. Thus, Hard-$\beta$, MaxVar-$\beta$,
and VADE-10\% acquire 90\% of the source pool as Selector-first prompts, of which 10\% are used
for training (Train) and 80\% are Probe-only, for 90\% total verified exposure. These methods are
useful static or online-selection references, but they are not verified-exposure matched to
strict-budget methods.

Strict-budget methods count every verified prompt toward the named budget, regardless of
acquisition route. In \method{}, REPR prompts initialize only the verified set and the selector
buffer, while the GRPO pool starts empty; verified REPR prompts enter the GRPO pool only through
later utility-based admission. Because the budget counts unique prompts, this admission adds no
exposure. All experiments use this configuration, and DOTS-MM uses the same accounting. For
\method{}-10\% on ViRL39K, the acquisition split is $1.16\%$ Selector-first ($450$ REPR prompts;
$K{=}150$, $R{=}3$) and $8.84\%$ Training-first, for exactly $10.00\%$ total verified exposure.
Depending on utility-based admission, Train lies in $[8.84\%,10.00\%]$ and Probe-only in
$[0.00\%,1.16\%]$; these are deterministic admission bounds, not confidence intervals. The observed
values are $9.57\%$ Train and $0.43\%$ Probe-only. For DOTS-MM-10\% on ViRL39K, the split is
$3.36\%$ Selector-first and $6.64\%$ Training-first, with Train $\in[6.64\%,10.00\%]$ (observed
$9.68\%$) and Probe-only $\in[0.00\%,3.36\%]$ (observed $0.32\%$). \method{} avoids dense verifier
scans: its selector overhead consists of representative probes plus non-generative embedding,
clustering, KNN transfer, and cluster allocation.

\section{Baseline configurations and validation}
\label{app:baseline-validation}
\paragraph{Baseline configurations.}
Random-$\beta$ selects a fixed random subset before training. Hard-$\beta$ ranks prompts by base-policy failure rate using one rollout per prompt over the full eligible training split, while MaxVar-$\beta$ ranks prompts by base-policy reward variance using $G$ rollouts per prompt.

DOTS-MM retains DOTS's scaled-attention transfer, target-difficulty selection, and 256-prompt reference set. Reference prompts are scored using the same vision-language policy and verifier used for GRPO and count toward the verified-prompt budget.

Both VADE variants retain independent per-prompt Beta posteriors and the published sampling and decay settings, initialized through a rollout scan of the full eligible training split. VADE-original has no training-pool cap, while VADE-$10\%$ limits the pool to 10\% of the full source pool.

Next we further elaborate on implementations of DOTS-MM and VADE. 

\paragraph{DOTS-MM.}
DOTS~\citep{sun2025dots} was introduced for text-only LLMs. Our multimodal
adaptation retains its scaled-attention transfer, target-difficulty selection,
and 256-prompt reference set. We varied the reference-set size and attention
temperature to assess the Spearman correlation between predicted and realized
utility. The highest correlation approached that of cosine-KNN with
$4{,}096$ reference prompts, exceeding the $10\%$ ViRL39K budget of $3{,}887$
prompts. These comparisons evaluate utility prediction; the benchmark
experiments use the original 256-prompt configuration of DOTS.

\paragraph{VADE.}
VADE~\citep{hu2025vade} maintains a separate Beta posterior for each prompt and
selects prompts through Thompson sampling. We evaluate two variants:
VADE-original, which has no training-pool cap, and VADE-$10\%$, which limits
the pool to $10\%$ of the full source pool. Both use the published sampling and
decay settings and initialize posteriors through a rollout scan of the eligible
training split. Each posterior incorporates its prompt's rollout outcomes,
while prior decay reduces the influence of earlier observations. In
VADE-$10\%$, $88.9\%$ of the eligible training split receives no new rollout
feedback after initialization.

\paragraph{Training stability.}
Neither adaptation showed optimization collapse during training.

\clearpage
\end{document}